\documentclass{article}
\pdfoutput=1
\usepackage{ijcai25}

\usepackage{times}
\usepackage{soul}
\usepackage{url}
\usepackage[hidelinks]{hyperref}
\usepackage[utf8]{inputenc}
\usepackage[small]{caption}
\usepackage{graphicx}
\usepackage{amsmath}
\usepackage{amsthm}
\usepackage{booktabs}
\usepackage{algorithm}
\usepackage{algorithmic}
\usepackage[switch]{lineno}
\usepackage{amssymb}
 \usepackage{multirow} 
 \usepackage{setspace}

\title{GBO: A Multi-Granularity Optimization Algorithm via Granular-ball for Continuous Problems}

\author{
Shuyin Xia$^{1,2,3}$,
Xinyu Lin$^{1,3}$,
Guan Wang$^{1,3}$,
De-Gang Chen$^{1,3}$,
Sen Zhao$^{1,2,3,4}$,
Guoyin Wang$^{1,4,5}$,
Jing Liang$^{6}$
\affiliations
\setstretch{0.7}
$^1$Chongqing Key Laboratory of Computational Intelligence\\
$^2$Key Laboratory of Big Data Intelligent Computing\\
$^3$Chongqing University of Posts and Telecommunications, Chongqing, China\\
$^4$Chongqing Normal University\\
$^5$National Center for Applied Mathematics in Chongging, Chongging Normal University, Chongging 401331, China\\
$^6$Zhengzhou University
\emails
xiasy@cqupt.edu.cn, 2022212013@stu.cqupt.edu.cn, 1640124287@qq.com,\\
chendegang0204@163.com, zhaosen@cqupt.edu.cn, wanggy@cqupt.edu.cn, liangjing@zzu.edu.cn\\
\setstretch{1}
}

\begin{document}

\maketitle

\begin{abstract}
    Optimization problems aim to find the optimal solution, which is becoming increasingly complex and difficult to solve. Traditional evolutionary optimization methods always overlook the granular characteristics of solution space. In the real scenario of numerous optimizations, the solution space is typically partitioned into sub-regions characterized by varying degree distributions. These sub-regions present different granularity characteristics at search potential and difficulty. Considering the granular characteristics of the solution space, the number of coarse-grained regions is smaller than the number of points, so the calculation is more efficient. On the other hand, coarse-grained characteristics are not easily affected by fine-grained sample points, so the calculation is more robust. To this end, this paper proposes a new multi-granularity evolutionary optimization method, namely the Granular-ball Optimization (GBO) algorithm, which characterizes and searches the solution space from coarse to fine. Specifically, using granular-balls instead of traditional points for optimization increases the diversity and robustness of the random search process. At the same time, the search range in different iteration processes is limited by the radius of granular-balls, covering the solution space from large to small. The mechanism of granular-ball splitting is applied to continuously split and evolve the large granular-balls into smaller ones for refining the solution space. Extensive experiments on commonly used benchmarks have shown that GBO outperforms popular and advanced evolutionary algorithms. The code can be found in the supporting materials.
\end{abstract}

\section{Introduction}
Optimization is a key research area in science and engineering, focused on identifying optimal solutions \cite{molaei2021particle}. It spans various fields, including engineering design \cite{liu2012extended,saha2021application,he2023high}, gene recognition \cite{xu2022inferring}, traffic signal control \cite{bi2014type,li2018signal}, machine learning \cite{barshandeh2022learning,abdollahzadeh2024puma,li2023survey}, and medical issues \cite{lian2024parrot}, among others.

Early works focused on deterministic search methods such as gradient descent \cite{tsitsiklis1986distributed,ruder2016overview}, Newton's method \cite{fischer1992special}, mixed integer programming \cite{shen2023advance}, etc. These methods usually require mathematical calculations and are prone to getting stuck in local optima. In large-scale environments, the solution space of optimization problems grows exponentially, making such methods no longer effective. 

The evolutionary optimization, inspired by natural evolution and biological behavior, has increasingly been applied to algorithm design and complex problem solving. Representative methods include genetic algorithm, particle swarm algorithm, ant colony algorithm \cite{holland1992genetic,kennedy1995particle,dorigo2006ant}, etc. These methods iteratively and randomly searches for the optimal solution through mutual learning and competition among individuals in the population. It does not rely on strict mathematical models, and can effectively handle complex optimization characteristics in big data environments.


Despite providing high-quality solutions to complex problems and attracting significant research interest, heuristic optimization methods often overlook the granular characteristics of different regions within the solution space. For instance, regions proximate to the global optimum are characterized by finer granularity, whereas regions farther from the optimum display coarser granularity. However, not all regions have the same optimal solution potential. 

Actually, modeling the granular characteristics of the solution space is not easy. Because there are several challenges: \textbf{(1)} The primary challenge is how to effectively characterize the granular characteristics of the solution space. Existing studies typically employ fine individual granularity to search the entire solution space, often neglecting its granular characteristics. This approach fails to address the complexity and diversity present in different regions effectively. 
\textbf{(2)} The second challenge lies in accurately assessing the potential optimality within each region. In the solution space, different regions may harbor varying degrees of optimal solutions, and traditional methods often struggle to precisely identify and evaluate these potentially optimal regions. Thus, it is crucial to develop an optimization algorithm capable of capturing the connections and differences between sub-regions of the solution space from a multi-granularity perspective.

To this end, we propose a multi-granularity optimization algorithm via granular-ball (GBO) for solving complex continuous optimization problems. Specifically, Multi-Granularity Solution Space Refinement involves initially covering the entire solution space with a coarse-grained granular-ball and then using a splitting mechanism to split fine-grained child granular-balls. Furthermore, Granular-ball Exploration and Exploitation involves the collaborative search among multiple child granular-balls. The coarse-to-fine search process better exploits potential differences in optimal solutions across various regions. These mechanisms replace the traditional point-based iterative search with a regional search approach, allowing for a more comprehensive consideration of the complexity and distinctiveness of the solution space. Experiments on benchmark and real-world problems show that GBO surpasses the classic and popular algorithms.

\textbf{In summary, our main contributions are listed as follows:}

\begin{itemize}
\item We examine the unique characteristics of the solution space using granular-balls, a novel approach for enhancing optimization performance in complex continuous problems.

\item We introduce a multi-granularity optimization algorithm termed Granular-Ball Optimization (GBO). This method innovatively models the solution space by transitioning from a coarse-grained to a fine-grained perspective through a two-stage process: \textbf{coarse-grained granular-ball initialization} and \textbf{fine-grained granular-ball splitting}. 

\item We designed a key granular-ball splitting strategy—\textbf{No-overlaping child granular-balls generation strategy}, which is the core of the granular-ball algorithm. This splitting strategy reduces the likelihood of the algorithm getting trapped in local optima and enhances the quality of the solutions found by the algorithm.

\item We validate the effectiveness of our proposed GBO method through extensive experiments conducted on both standard benchmark problems \cite{liang2013problem} and challenging real-world optimization scenarios. 
\end{itemize}

\section{Related Work}

\subsection{Evolutionary Algorithms and the Fireworks Algorithm}

Evolutionary Algorithms (EAs) are a class of global optimization algorithms inspired by natural evolution. They employ population-based strategies that iteratively improve solutions through mechanisms like selection, crossover, and mutation. Examples include Genetic Algorithms (GAs) \cite{holland1992genetic}, Evolution Strategies (ES) \cite{beyer2002evolution}, Ant Colony Optimization (ACO) \cite{dorigo2006ant}, and Particle Swarm Optimization (PSO) \cite{kennedy1995particle}.

EAs have also been successfully applied to various real-world problems, with recent advancements focusing on improving convergence speed and addressing challenges like local optima. For example, \cite{xiang2019simple} proposed a PSO strategy (PBS-PSO) that utilizes PID control to accelerate convergence, while \cite{zhang2018competitive} introduced a competitive mechanism to enhance the exploration capabilities of PSO.

The Fireworks Algorithm (FWA), proposed in 2010 \cite{tan2010fireworks}, is another nature-inspired optimization algorithm that simulates the explosive behavior of fireworks to explore the search space \cite{soares2024synchronisation}. Despite its potential, the original FWA has limitations, prompting various enhancements. These include the Enhanced Fireworks Algorithm (EFWA) \cite{pekdemir2024efficient}, which optimizes core components, and adaptive versions like AFWA and dynFWA \cite{zheng2014dynamic} that dynamically adjust explosion amplitudes. Hybrid methods, such as CoFFWA, improve information exchange between fireworks, while GPU-based implementations enhance performance on large-scale problems.

Further advancements, such as LoTFWA \cite{li2017loser} and MGFWA \cite{meng2024multi}, have incorporated competitive mechanisms and multi-guiding sparks to address challenges like local stagnation and improve performance on multi-modal problems. These refinements demonstrate the ongoing evolution of FWA and its potential as a valuable tool in the field of swarm intelligence.

\subsection{Granular-ball Computing (GBC)}
 Chen et al.\cite{chen1982topological} pointed out that human cognition has the law of ``global precedence” in his research published in Science. Based on the theoretical basis of traditional granular computing, Wang et al.\cite{wang2017dgcc} took the lead in proposing multi-granular cognitive computing in combination with the cognitive law in human brain cognition. Xia et al.\cite{xia2023granular} introduced an innovative computational method known as granular-ball computing (GBC), celebrated for its efficiency and robustness. 

The reason for Xia et al.\cite{xia2023granular}'s approach to multi-granularity feature representation using granular-ball is that the geometry of a granular-ball is completely symmetric and has the most concise, standard mathematical representation. Therefore, it facilitates expansion into higher dimensional space. Compared with the traditional method which takes the most fine-grained points as input, the granular-ball computing takes the coarse-grained granular-balls as input, which is efficient, robust, and interpretable \cite{xia2023granular}. Granular-ball computing has been extensively applied across diverse domains within artificial intelligence, as demonstrated by studies such as those by Xie et al., Qadir et al., Zhang et al., and Liu et al.\cite{xie2024gbg++,quadir2024granular,zhang2023incremental,liuunlock}. However, its application in optimization is relatively under-explored. Thus, this paper proposes a multi-granularity granular-ball optimization algorithm to explore this domain.

\section{The Proposed Algorithm}

\begin{figure*}[t]
\centering
\includegraphics[width=1.9\columnwidth]{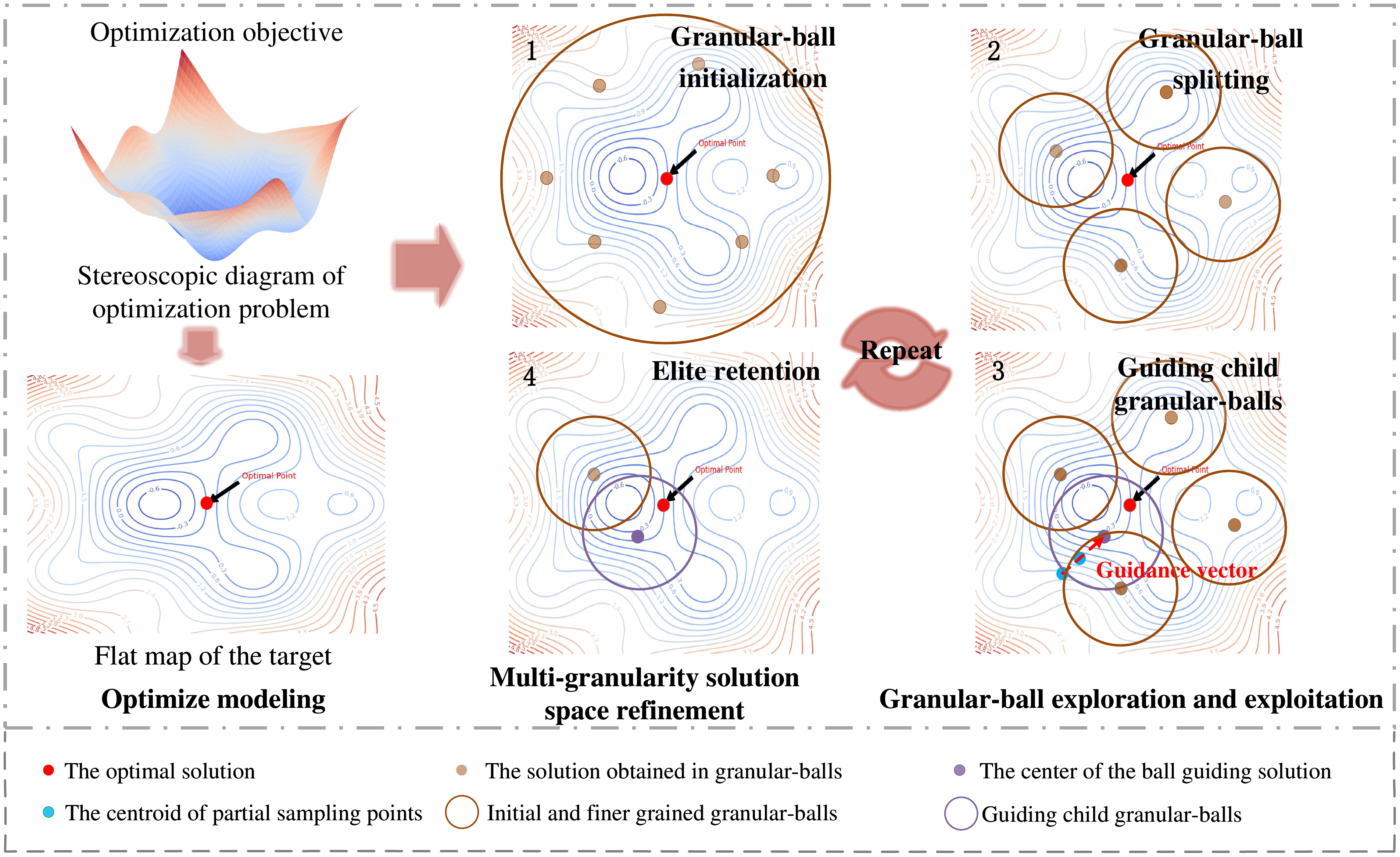} 
\caption{The framework of the proposed GBO. The figure illustrates the process of GBO using multi-granularity solution space refinement and granular-ball exploration and exploitation to solve optimization problems.} 
\label{fig2}
\end{figure*}

In this section, we present the multi-granularity optimization algorithm via granular-ball (GBO) for solving optimization problems (shown in Figure \ref{fig2}),  which is composed of two modules: (1) Multi-Granularity Solution.
Space Refinement: the solution space is refined from coarse-grained and fine-grained perspectives, respectively; (2) Granular-ball Exploration and Exploitation: the optimal solution is found through cooperative search among child granular-balls.

\subsection{Multi-Granularity Solution Space Refinement}

In this module, based on the ``global precedence" cognitive law \cite{chen1982topological}, a coarse-grained initial granular-ball is used to cover the solution space of the objective function. Then the sampling points operation is carried out inside the initial granular-ball to split many child granular-balls to refine the solution space.

\noindent \textbf{Coarse-Grained Initialization.}In any dimension, a granular-ball needs only two data points to represent it: the center $c$ and the radius $r$. 
In a space of arbitrary dimensions, a granular-ball ($\mathcal{GB}$) is defined by its center vector $c$ and radius vector $r$. Given an initial granular-ball $\mathcal{GB}$, the center $c$ represents the position of the granular-ball in space and is a vector pointing to the center of the solution space. The radius $r$ is a vector where each component corresponds to half of the range in the respective dimension of the solution space. The initial granular-ball covers the entire solution space to ensure that no potential optimal solutions are overlooked.


The fitness value, as the only solution quality evaluation indicator in evolutionary computation, is indispensable for the algorithm. In this paper, due to the use of granular-balls instead of points to evaluate the search potential of a certain region in the solution space during the algorithm optimization process, the fitness of the search individuals in the algorithm, that is, the quality of the granular-balls, is redefined. The fitness value of the center of a granular-ball is taken as the quality of the granular-ball:
    
\begin{equation}
quality(\mathcal{GB})=f(c).
\end{equation}

As shown in 1 in Figure \ref{fig2}, the process begins with a large granular-ball covering the solution space. This approach ensures that no potential optimal solution location is overlooked while also enabling partitioning of the entire space. 

\noindent \textbf{Fine-Grained Splitting.} When each granular-ball splits, the radius of the granular-ball is gradually reduced, and this process is also a transition from coarse-grained to fine-grained exploration. This strategy makes up for the shortcoming of the traditional evolutionary method that can only search on a single granularity and greatly improves the robustness of the algorithm to deal with problems of different complexity. In other words, the radii of the parent granular-ball and child granular-balls meet the following formula:

\begin{equation}
r_{t+1}=r_t\times \rho, t=1,2,..,t_{max}-1
\end{equation}

\noindent where $r_{t+1}$ denotes the radii of parent granular-balls in $t+1$-th iteration, $r_t$ denotes the radius of parent granular-ball in $t$-th iteration, $\rho$ denotes the rate of radius reduction.

As depicted in 4 in Figure \ref{fig2}, after the splitting process, only the promising child granular-balls are retained. Compared to their parent granular-balls, these child granular-balls conduct a more fine-grained search within the solution space. Subsequently, each child granular-ball becomes a new parent granular-ball, resembling the process of using a microscope. Initially, a lower magnification is used to locate the area of interest, and then the position is gradually refined and magnified until the target is identified.

\subsection{Granular-ball Exploration and Exploitation}

Starting from an initial granular-ball that covers the solution space, each generation of granular-balls will undergo three processes: sampling points within the granular-ball, splitting, and selection. The radius of the granular-balls will gradually decrease, allowing for a more refined search of the solution space. The overall framework of the GBO algorithm is presented in Algorithm \ref{alg1}.



\noindent \textbf{Sampling points within the granular-ball.}
The process for generating \( n^*\) sampling points within a granular ball is as follows: First, \( n^* \times D \) uniform random numbers are generated in the interval \([0, 1]\), satisfying the condition \( rand_k^{j} \sim \text{U}(0, 1) \), where \( k = 1, 2, \ldots, n^* \) and \( j = 1, 2, \ldots, D \). Here, \( rand_k^{j} \) denotes the \( k \)-th random number in dimension \( j \), and \( D \) represents the dimension of the granular ball.

Then, the position of $k$-th sample point in dimension $j$ in the  granular-ball is calculated based on random numbers:

\begin{equation}
x_{k}^{j}=lb_j+rand_k^{j} \times (ub_j-lb_j), 
\end{equation}

\noindent Where $k = 1, 2, \dots, n^*$ and $j = 1, 2, \dots, D$, $lb_j$ is the lower bound of $\mathcal{GB}$ in the $j$-th dimension, and $ub_j$ is the upper bound of $\mathcal{GB}$ in the $j$-th dimension. After determining the sampling points, any $x_k^j$ that falls outside the defined range is randomly remapped back into the specified domain.

\begin{algorithm}[!t]
    \renewcommand{\algorithmicrequire}{\textbf{Input:}}
    \renewcommand{\algorithmicensure}{\textbf{Output:}}
    \caption{No-overlaping child granular-balls generation strategy}
    \label{alg2} 
    \begin{algorithmic}[1]
        \REQUIRE Sampling point set $S$ and granular-ball  $\mathcal{GB}$.
        \ENSURE The non-overlapping child granular-balls $C_1$ in $\mathcal{GB}$.
        \STATE $C_1 \gets \left\{  \right\} $;
        \STATE Obtain the radius $r$ of granular-ball  $\mathcal{GB}$;
        \STATE $r \gets r \times  \rho $
        \FOR{$bp$ \textbf{in} $S$}
            \IF{\( bp \) is not within any granular-ball in \( C_1 \)}
             \STATE Generate a granular-ball centered at \( bp \) with radius \( r \) and add it to \( C_1 \);
            \ENDIF
        \ENDFOR
    \RETURN $C_1$;
    \end{algorithmic}  
\end{algorithm}

Determining the number of sampling points for each particle sphere in each generation is a crucial process. The sampling points strategy can be mathematically expressed as

\begin{equation}
n = \frac{fes_{max}}{t_{max}},
\end{equation}

\noindent where \( n \) represents the total number of sampling points in each iteration, \( fes_{max} \) indicates the maximum number of fitness evaluations, and \( t_{max} \) denotes the maximum number of iterations. 

Thus, in each iteration, the number of sampling points for each granular ball, denoted as \( \tilde{n} \), must satisfy the condition \( \tilde{n} = \frac{n}{|G|} \), where \( |G| \) represents the number of granular balls in that generation. This sampling strategy enhances the algorithm's adaptability across various problems.

Specifically, $\tilde{n}_1$ sampling points are first generated randomly for each parent granular-ball. Then, some child granular-balls are generated according to the no-overlapping generation strategy to maintain the diversity of granular-balls. Then, based on $\tilde{n}_1$ sampling points, $\tilde{n}_2$ guiding points are generated using the idea of gradient descent, and some guiding child granular-balls are generated at these points. $\tilde{n}_1$ and $\tilde{n}_2$ satisfy $\tilde{n} = \tilde{n}_1 + \tilde{n}_2$.


\noindent \textbf{No-overlaping child granular-balls generation strategy.}
In this strategy, to maintain better exploration of a parent granular-ball, we aim for the child granular-balls formed by its splitting to be non-overlapping. Specifically, the centers of child granular-balls originating from the same parent granular-ball should not fall within the volume of another child granular-ball from that parent. As shown in 2 in Figure \ref{fig2}, the center of each child granular-ball does not overlap with the center of another child granular-ball. This arrangement allows the child granular-balls to better partition the interior of the parent granular-ball, dividing the original granularity into finer levels, which facilitates a more detailed search. The framework of this part of the algorithm is presented in Algorithm \ref{alg2}. For a parent granular-ball, initialize a set of child granular-balls $C_1$. Each time, randomly select a sampling point from the set of $\tilde{n}_1$ sampling points as the set \( S \). If the sampling point is not inside any of the child granular-balls generated by the parent granular-ball, i.e., it satisfies the condition for all child granular-balls in the set $C_1$:

\begin{equation}
\lvert x_{k}-c_i \rvert \geq r_{t+1}, i=1,2,...,\lvert C_1 \vert
\end{equation}

\noindent where $x_{k}$ denotes the $k$-th sampling points, $c_{i}$ denotes the center of the $i$-th granular-ball, $\lvert x_{k}-c_i \rvert$ denotes the Euclidean distance between $x_{k}$ and $c_{i}$. This formula indicates that the center of a generated child granular-ball should not be inside the previously generated child granular-balls. If this condition is satisfied, a child granular-ball is generated with the sampling point as its center and a radius of $r_{t+1}$, and it is added to the set $C_1$.

\begin{algorithm}[!t]
    \renewcommand{\algorithmicrequire}{\textbf{Input:}}
    \renewcommand{\algorithmicensure}{\textbf{Output:}}
    \caption{Guiding child granular-balls generation strategy}
    \label{alg3}
    \begin{algorithmic}[1]
        \REQUIRE Sampling point set $S$ and granular-ball  $\mathcal{GB}$.
        \ENSURE The guiding child granular-balls  in  $\mathcal{GB}$.
        \STATE Obtain the radius $r$ of granular-ball  $\mathcal{GB}$;
        \STATE Sort the fitness values of the sampling points in \( S \) in ascending order;
        \STATE $C_2 \gets \left\{  \right\} $;
        \STATE $r \gets r \times  \rho $;
        \STATE $c^{t}\gets \frac{1}{\tilde{n}_1\times \sigma}\sum_{j=1}^{\tilde{n}_1\times \sigma}{f}\left( s_{j} \right)$ ;
        \STATE $c^{b}\gets \frac{1}{\tilde{n}_1\times \sigma}\sum_{j=\tilde{n}_1-\tilde{n}_1\times \sigma +1}^{\tilde{n}_1}{f}\left( s_{j} \right) $;
        \STATE  ${\Delta}\gets c^{t}-c^{b}$;
        \FOR{$i = 1$ \TO $\tilde{n}_2$}
            \STATE Sample w from a specific distribution;
            \STATE $\tilde{c}\gets c^{t}+ {\Delta}\times {w}$;
            \STATE Generate a granular-ball centered at $\tilde{c}$ with radius \( r \) and add it to \( C_2 \);
        \ENDFOR
    \RETURN $C_2$
    \end{algorithmic}  
    
\end{algorithm}

\noindent \textbf{Guiding child granular-balls generation strategy.} 
In Section 2, we introduced the evolution of the Fireworks Algorithm (FWA). During this process, GFWA \cite{7508443} introduced the use of gradient information from sampled points within the explosion of the fireworks, enhancing the algorithm's optimization capability by adding guiding sparks. This method has been preserved and continuously optimized in subsequent FWAs \cite{li2017loser}, \cite{meng2024multi}. In the ablation study, even when only the first child granular-ball generation method was employed, GBO’s performance on the test set already surpassed MGFWA (the state-of-the-art variant of FWA) \cite{meng2024multi}. However, due to certain geometric similarities between granular-balls and fireworks, we drew inspiration from GFWA’s approach to further improve the performance of the granular-ball algorithm, incorporating gradient information to generate guiding granular-balls.
The process of calculating the guiding vector to generate child granular-balls can be described as follows(Algorithm 2). Firstly, sort the fitness corresponding to the $\tilde{n}_1$ sampling points in ascending order. Secondly, select the top and bottom groups based on these sampling points. Calculate the centroids of two sets of sampling points as follows:

\begin{equation}
c_{i}^{t}=\frac{\sum_{j=1}^{\tilde{n_1} \times \sigma} f(s_{j})}{\tilde{n_1} \times \sigma},
\end{equation}

\begin{equation}
c_{i}^{b}=\frac{\sum_{j=\tilde{n_1}-\tilde{n_1} \times \sigma+1}^{\tilde{n_1}} f(s_{j})}{\tilde{n_1} \times \sigma},
\end{equation}

\noindent where $s_{j}$ is the sampling point in \( S \) with the \( j \)-th fitness value after sorting, $f(s_{j})$ denotes the fitness of $s_{j}$, $\sigma$ is a hyper-parameter to control the number of sampling points in each group, $c_{i}^{t}$ and $c_{i}^{b}$ are the centroids of the two groups by the i-th granular-ball. Then, the guiding vector $ {\Delta}_{i}$ is estimated by the difference between the two centroids in the i-th granular-ball:

\begin{equation}
 {\Delta}_{i}=c_{i}^{t}-c_{i}^{b}.
\end{equation}

Subsequently, the central position of $\tilde{n}_2$ guiding granular-balls are given:

\begin{equation}
\tilde{c}=c_{i}^{t}+ {\Delta}_{i}\times {w_i},
\end{equation}

\noindent where $\tilde{c}$ denotes the center of $\tilde{n}_2$ guiding granular-balls,
${w}$is the weight that controls the length of the guiding vector, and it satisfies a random uniform distribution in the interval [0.5, 1.5]. The guiding granular-ball strategy further improves the convergence speed of GBO (3 in Figure \ref{fig2}). 

\noindent \textbf{Elite retention.} Typically, a generation of parent granular-balls produces many child granular-balls, which will waste lots of computational resources if they are all retained for the next iteration. Therefore, if the number of balls exceeds $N$, we sort the quality of all child granular-balls and select $N$ elite child granular-balls as the new generation granular-ball population for iterative search, otherwise all reserved.

\noindent \textbf{Iteration Loop.} The above multi-granularity design mechanisms for solution space and search methods work closely together to help GBO effectively find the optimal solution from coarse to fine granularity, making the algorithm capable of solving different optimization problems. Usually, after splitting to produce a new generation of granules, a new round of search will be conducted with them as the main body, and the search will be iteratively repeated until the consumption of computing resources is completed.

\begin{algorithm}[!t]
    \renewcommand{\algorithmicrequire}{\textbf{Input:}}
    \renewcommand{\algorithmicensure}{\textbf{Output:}}
    \caption{The multi-granularity optimization algorithm via granular-ball (GBO)}
    \label{alg1}
    \begin{algorithmic}[1]
        \REQUIRE The optimization objective $f$ and maximum number of iterations $fes_{max}$.
        \ENSURE The best fitness of $f^*$ and its corresponding solution position ${bp}^*$.
        \STATE $G \gets \left\{  \right\} $;
        \STATE $n \gets \frac{fes_{max}}{t_{max}}$;
        \STATE Initialize a granular-ball that covers the solution space and add it to $G$;
        \FOR{$j= 1$ \TO $t_{\max}$}
            \STATE $\tilde{n} \gets \frac{n}{\left| G \right|}$;
            \STATE $\tilde{n}_1\gets $  $\tilde{n}$ - $\tilde{n}_2$;
            \STATE $G_{child} \gets  \left\{  \right\}$ ;
            \STATE Calculate the fitness values of the sampling point set $S$;
            \FOR{$i= 1$ \TO $\left| G \right|$}
                \STATE Generate $\tilde{n}_1$ \text{sampling points  within } ${\mathcal{GB}}_i$ as $S_i$;
                \STATE Perform random mapping on $S_i$;
                \STATE $C_1 \gets \text{Alg.\ref{alg2}} ($$S_i$$, {\mathcal{GB}}_i)$;
                \STATE $C_2 \gets \text{Alg.\ref{alg3}} ($$S_i$$, {\mathcal{GB}}_i)$;
                \STATE $G_{child} \gets G_{child} \cup C_1 \cup C_2$;
            \ENDFOR
        \STATE Sort the child granular-balls in $G_{child}$ in ascending order of mass;
        \STATE \text{Select} $\min \left\{ \left| G \right|,N \right\} $ \text{elite granular-balls as} $G$ \text{in} 
        $G_{child}$;
        \STATE Update $f^*$ and ${bp}^*$;
        \ENDFOR
    \RETURN $f^*$ and ${bp}^*$;
    \end{algorithmic}  
\end{algorithm}

\section{Experiments}
\subsection{Experiment Settings}

\noindent \textbf{Benchmarks.} To verify the effectiveness of the GBO proposed in this paper, we conduct experiments on a commonly used CEC2013 benchmark \cite{liang2013problem}. There are 28 evaluation functions in the CEC2013 benchmark, including 5 unimodal functions, 15 basic multimodal functions, and 8 composition functions. In addition, to validate the effectiveness of GBO in solving practical optimization problems, we conducted experiments using the Spread Spectrum Radar Polly Phase Code Design problem \cite{das2010problem}.
For a fair comparison, the number of given fitness evaluations for all algorithms is set to $10000\times D$. This paper provides the mean errors (Mean) and standard deviations (Std.) obtained from 51 independent runs to assess the performance of all methods. Meanwhile, the specific experimental setup for GBO is: $\rho=0.96$, $N=30$, $t_{max}=250$, $\sigma=0.2$ \cite{meng2024multi}, $\tilde{n}_2=2$. We mainly presented the results of all algorithms in 30 dimensions for illustration purposes. In addition, for strict comparison, the Wilcoxon rank sum test was used at the significance level of $\alpha=0.05$. Moreover, at a significance level of $\alpha=0.05$, the Friedman test was used to comprehensively analyze each method's average rank (AR) on an overall problem set.


 \begin{table*}[ht]
			\setlength{\abovecaptionskip}{0.3cm}
			\setlength{\belowcaptionskip}{0.3cm}
			\scriptsize
			\centering
           \caption{Comparison of GBO with several popular variants of single objective global optimization algorithms in $\textbf{30}$-$D$.}
        \setlength{\tabcolsep}{0.2mm}
            \fontsize{7}{10}\selectfont
            \resizebox{\textwidth}{!}{ 
        \begin{tabular}{|c|cc|cc|cc|cc|cc|cc|cc|}
            \hline
            \multirow{2}{*}{\textit{f}} & \multicolumn{2}{c|}{GBO}             & \multicolumn{2}{c|}{JADE}              & \multicolumn{2}{c|}{MGFWA}             & \multicolumn{2}{c|}{NSHADE}              & \multicolumn{2}{c|}{LSHADE}           & \multicolumn{2}{c|}{PVADE}           & \multicolumn{2}{c|}{SPSO2011}  \\ \cline{2-15} 
            & \multicolumn{1}{c|}{Mean} & Std.     & \multicolumn{1}{c|}{Mean} & Std.     & \multicolumn{1}{c|}{Mean} & Std.     & \multicolumn{1}{c|}{Mean} & Std.     & \multicolumn{1}{c|}{Mean} & Std.     & \multicolumn{1}{c|}{Mean} & Std.   & \multicolumn{1}{c|}{Mean} & Std.  \\ \hline
            $F_{1}$                          & 3.25E-05	& 5.79E-06	&  \textbf{0.00E+00}- 	& \textbf{0.00E+00} 	& 3.57E-14- 	& 8.27E-14 	& 2.23E-13- 	& 3.15E-14 	& \textbf{0.00E+00}- 	& \textbf{0.00E+00} 	& \textbf{0.00E+00}- 	& \textbf{0.00E+00} 	& 8.92E-14- 	& 1.11E-13  \\
            $F_{2}$                          & 8.30E+05	&4.46E+05	&  \textbf{7.85E+03}- 	& \textbf{6.02E+03} 	&1.41E+06+	&4.95E+05	& 4.86E+04- 	& 2.97E+04 	& 1.16E+04- 	& 8.62E+03 	&2.12E+06+	&1.54E+06	& 2.31E+05 -	& 8.80E+04 \\
            $F_{3}$                          & \textbf{3.37E+01}	&\textbf{4.37E+00}	& 4.91E+05+	&2.09E+06	&6.42E+06+	&9.46E+06	&3.05E+06+	&1.55E+07	&7.62E+05+	&2.14E+06	&1.65E+03+	&2.80E+03	&1.89E+07+	&1.97E+07\\
            $F_{4}$                          & 3.60E+04	&9.53E+03	&  3.44E+03- 	& 1.82E+03 	& 1.22E+03- 	& 4.43E+02 	& 2.18E+04- 	& 3.26E+04 	& \textbf{2.03E-04}- 	& \textbf{4.49E-04} 	& 1.70E+04- 	& 2.82E+03 	& 6.67E+03- 	& 1.67E+03 \\
            $F_{5}$                          & 3.53E-03 & 4.12E-04	&  \textbf{1.09E-13}- 	& \textbf{2.21E-14} 	&6.57E-03+	&1.85E-03	& 2.76E-13- 	& 9.35E-14 	& 1.14E-13- 	& 1.26E-29 	& 1.40E-07- 	& 1.84E-07 	& 9.34E-04- 	& 9.02E-05 \\
            $F_{6}$                          & 2.08E+01	&1.89E+01	&  \textbf{2.07E+00}- 	& \textbf{7.10E+00} 	& 1.49E+01- 	& 2.12E-01 	& 6.05E+00- 	& 4.34E+00 	& 2.77E+00- 	& 6.97E+00 	& 8.29E+00- 	& 5.76E+00 	&2.13E+01+	&2.18E+01\\
            $F_{7}$                          & \textbf{6.29E-02}	&\textbf{8.95E-02}	& 4.36E+00+	&4.80E+00	&2.56E+01+	&8.42E+00	&5.98E+01+	&1.51E+01	&4.84E+00+	&4.55E+00	&1.29E+00+	&1.20E+00	&1.82E+01+	&9.34E+00\\
            $F_{8}$                          & 2.09E+01	&5.81E-02	& 2.09E+01$\approx$	&5.07E-02	& \textbf{2.08E+01}- 	& \textbf{5.94E-02} 	&2.09E+01$\approx$	&5.26E-02	&2.09E+01$\approx$	&5.51E-02	&2.09E+01$\approx$	&4.77E-02	&2.09E+01$\approx$	&7.00E-02\\
            $F_{9}$                          & \textbf{3.66E+00}	&\textbf{1.93E+00}	& 3.24E+01+	&1.40E+00	&9.98E+00+	&1.82E+00	&2.90E+01+	&1.43E+00	&2.77E+01+	&1.84E+00	&6.30E+00+	&3.24E+00	&2.60E+01+	&5.07E+00\\
            $F_{10}$                         & \textbf{1.66E-03}	&\textbf{2.72E-03}	& 3.30E-02+	&1.73E-02	&2.53E-02+	&2.00E-02	&5.91E-02+	&4.84E-02	&7.60E-02+	&5.36E-02	&2.16E-02+	&1.34E-02	&1.96E-01+	&8.93E-02\\
            $F_{11}$                         & 2.10E+01	&4.96E+00	&  \textbf{0.00E+00}- 	& \textbf{0.00E+00} 	&2.54E+01+	&5.40E+00	& 5.80E-14- 	& 1.37E-14 	& \textbf{0.00E+00}- 	& \textbf{0.00E+00} 	&5.84E+01+	&1.10E+01	&5.43E+01+	&2.73E+01\\
            $F_{12}$                         & \textbf{2.13E+01}	&\textbf{4.99E+00}	& 5.16E+01+	&1.45E+01	&2.65E+01+	&5.70E+00	&4.73E+01+	&1.00E+01	&2.42E+01+	&3.26E+00	&1.15E+02+	&1.13E+01	&4.11E+01+	&1.21E+01\\
            $F_{13}$                         & \textbf{3.85E+01}	&\textbf{1.25E+01}	& 7.01E+01+	&1.55E+01	&5.60E+01+	&1.31E+01	&1.04E+02+	&1.92E+01	&4.79E+01+	&9.99E+00	&1.31E+02+	&1.23E+01	&8.91E+01+	&1.92E+01\\
            $F_{14}$                         & 7.73E+02	&2.74E+02	&  5.10E-02- 	& 2.87E-02 	&2.39E+03+	&3.58E+02	& 4.36E+00- 	& 1.41E+00 	& \textbf{4.57E-02}- 	& \textbf{2.97E-02} 	&3.20E+03+	&4.34E+02	&4.82E+03+	&5.94E+02\\
            $F_{15}$                         & \textbf{7.88E+02}	&\textbf{2.56E+02}	& 6.54E+03+	&3.86E+02	&2.29E+03+	&3.25E+02	&3.17E+03+	&3.46E+02	&3.44E+03+	&3.27E+02	&5.61E+03+	&3.15E+02	&4.30E+03+	&4.18E+02\\
            $F_{16}$                         & \textbf{6.71E-03}	&\textbf{2.57E-03}	& 2.37E+00+	&2.82E-01	&4.97E-02+	&1.32E-02	&8.00E-01+	&1.48E-01	&1.12E+00+	&1.74E-01	&2.39E+00+	&2.63E-01	&1.39E+00+	&2.80E-01\\
            $F_{17}$                         & 4.84E+01	&3.93E+00	&  \textbf{3.04E+01}- 	& \textbf{0.00E+00} 	&5.60E+01+	&4.88E+00	& 3.05E+01- 	& 3.01E-02 	& \textbf{3.04E+01}- 	& \textbf{2.63E-14 }	&1.02E+02+	&1.16E+01	&1.28E+02+	&2.33E+01\\
            $F_{18}$                         & \textbf{4.78E+01}	&\textbf{3.65E+00}	& 1.70E+02+	&9.47E+00	&5.65E+01+	&5.44E+00	&8.75E+01+	&8.04E+00	&7.80E+01+	&5.75E+00	&1.82E+02+	&1.19E+01	&1.09E+02+	&9.87E+00\\
            $F_{19}$                         & 3.00E+00	&5.43E-01	& 3.50E+00+	&3.71E-01	& 2.39E+00- 	& 4.13E-01 	& 1.84E+00- 	& 5.83E-01 	& \textbf{1.46E+00}- 	& \textbf{1.21E-01} 	&5.40E+00+	&8.02E-01	&5.66E+00+	&2.93E+00\\
            $F_{20}$                         & \textbf{8.68E+00}	&\textbf{6.81E-01}	& 1.18E+01+	&2.83E-01	&1.27E+01+	&1.29E+00	&1.50E+01+	&2.22E-01	&1.11E+01+	&3.84E-01	&1.13E+01+	&3.24E-01	&1.07E+01+	&5.75E-01\\
            $F_{21}$                         & 2.48E+02	&9.67E+01	& 2.83E+02$\approx$	&5.89E+01	& \textbf{2.11E+02}- 	& \textbf{3.00E+01} 	&3.12E+02+	&7.38E+01	&2.98E+02+	&5.99E+01	&3.19E+02+	&6.20E+01	&3.19E+02+	&5.73E+01\\
            $F_{22}$                         & 8.62E+02	&2.21E+02	&  2.01E+02- 	& 2.40E+02 	&2.78E+03+	&4.07E+02	& \textbf{9.23E+01}- 	& \textbf{2.90E+01} 	& 1.06E+02- 	& 1.29E+01 	&2.50E+03+	&3.82E+02	&3.97E+03+	&6.60E+02\\
            $F_{23}$                         & \textbf{8.57E+02}	&\textbf{2.66E+02}	& 6.51E+03+	&3.93E+02	&2.93E+03+	&4.76E+02	&3.98E+03+	&3.74E+02	&3.74E+03+	&4.16E+02	&5.81E+03+	&4.99E+02	&4.21E+03+	&5.83E+02\\
            $F_{24}$                         & \textbf{2.00E+02}	&\textbf{1.45E-02}	& 2.42E+02+	&2.40E+01	&2.03E+02+	&2.46E+00	&2.29E+02+	&9.82E+00	&2.16E+02+	&1.37E+01	&2.02E+02+	&1.38E+00	&2.28E+02+	&6.79E+00\\
            $F_{25}$                         & \textbf{2.12E+02}	&\textbf{1.94E+01}	& 2.85E+02+	&7.81E+00	&2.47E+02+	&1.31E+01	&2.91E+02+	&1.87E+01	&2.83E+02+	&4.32E+00	&2.30E+02+	&2.06E+01	&2.65E+02+	&6.66E+00\\
            $F_{26}$                         & \textbf{2.00E+02}	&\textbf{2.56E-02}	& 2.35E+02+	&6.37E+01	&\textbf{2.00E+02}$\approx$	&\textbf{1.48E-02}	&\textbf{2.00E+02}$\approx$	&\textbf{3.52E-01}	&2.06E+02+	&2.91E+01	&2.18E+02$\approx$	&3.97E+01	&2.17E+02+	&4.38E+01\\
            $F_{27}$                         & \textbf{3.03E+02}	&\textbf{1.80E-01}	& 9.26E+02+	&1.98E+02	&3.44E+02+	&2.96E+01	&8.60E+02+	&1.22E+02	&8.70E+02+	&1.17E+02	&3.26E+02+	&1.13E+01	&5.80E+02+	&5.55E+01\\
            $F_{28}$                         & 3.00E+02	&1.25E-02	& 3.00E+02$\approx$	&2.26E-13	& \textbf{2.96E+02}- 	& \textbf{2.77E+01} 	& \textbf{2.96E+02}- 	& \textbf{2.77E+01} 	&3.00E+02$\approx$	&2.03E-13	&3.00E+02$\approx$	&3.22E-05	& \textbf{2.96E+02}- 	& \textbf{2.77E+01} \\ \hline
            \textit{w/t/l}              & \multicolumn{2}{c|}{-}               & \multicolumn{2}{c|}{16/3/9}          & \multicolumn{2}{c|}{20/1/7}          & \multicolumn{2}{c|}{15/2/11}          & \multicolumn{2}{c|}{16/2/10}          & \multicolumn{2}{c|}{21/3/4}         & \multicolumn{2}{c|}{22/1/5} \\ \hline
            \textit{Rank}               & \multicolumn{2}{c|}{\textbf{2.82}}            & \multicolumn{2}{c|}{4.25}            & \multicolumn{2}{c|}{3.61}        & \multicolumn{2}{c|}{4.27}            & \multicolumn{2}{c|}{3.18}            & \multicolumn{2}{c|}{4.73}            &\multicolumn{2}{c|}{5.14} \\ \hline
        \end{tabular}}
        \label{accuracy_30D_2}
    \end{table*}

\noindent \textbf{Comparison Methods.} This paper first conducted a comprehensive comparison with classic evolutionary algorithms, including PSO \cite{kennedy1995particle}, DE \cite{qin2008differential}, GA \cite{holland1992genetic}, ABC \cite{karaboga2014comprehensive}, SHADE \cite{tanabe2013success} and LoTFWA \cite{li2017loser}. Subsequently, GBO is compared with several popular variants of single objective global optimization algorithms including JADE \cite{zhang2009jade}, MGFWA \cite{meng2024multi} (the SOTA variant of FWA), NSHADE \cite{ghosh2022using}, LSHADE \cite{tanabe2014improving} (CEC 2014's champion algorithm), PVADE \cite{dos2013population} and SPSO2011 \cite{zambrano2013standard} to further verify the performance of GBO. The algorithm parameters are shown in Table \ref{table2} (Appendix 6.3).

\noindent \textbf{Overall Performance.} The experimental results are shown in Table \ref{accuracy_30D_2} and Table \ref{accuracy_30D} (Appendix 6.3). For each function, the optimal result is displayed in bold for emphasis. The mean errors followed by “+" indicate that GBO has good performance, the errors followed by “-" indicate that the comparison method has good performance, and the errors followed by ``$\approx$" indicate that the performance of GBO and comparison method is similar. In the comparison experiments with classic algorithms, the performance of GBO exceeded that of the classic comparison algorithms by 61\%, 64\%, 75\%, 93\%, 54\%, and 71\%, respectively. In addition, GBO has a mean rank of 2.52 across the 28 functions, which is far better than that of the comparison classic algorithms. As can be seen from Table \ref{accuracy_30D_2}, the performance of GBO is 57\%, 71\%, 54\%, 57\%, 75\%, and 79\% above the other six algorithms, respectively. In addition, the AR of GBO in 28 functions is 2.82, which is better than the comparison algorithms. The algorithm performs significantly better in testing complex functions compared to simpler ones, mainly due to the independent search between different granular-balls, resulting in good diversity.

\noindent \textbf{Ablation Studies.}
 We performed ablation experiments on the CEC2013 benchmark to examine the effects of the strategies described in the previous section on GBO. It mainly includes GBO-w/o guiding granular-balls. The results show that when GBO does not use guiding strategy, 1 of the 28 test functions are better than GBO, 15 functions are equal to GBO, and 12 function is worse than GBO.

This indicates that the guiding granular-ball strategy has played a significant role in assisting the model to solve optimization problems, thereby improving the efficiency of the algorithm. This is because the centroids of mass guiding the granular-balls effectively dictate the subsequent search directions for the elite granular-balls within the solution space by aggregating information from high-quality sampling points. This mechanism not only enhances the efficiency of the search but also ensures a more precise approximation of the global optimal solution. Consequently, GBO is capable of rapidly identifying potentially favorable areas within a complex solution space and conducting thorough explorations therein, thereby significantly improving both the quality of insights and the precision of the search.

\begin{figure}[t]
\centering
\includegraphics[width=1\columnwidth]{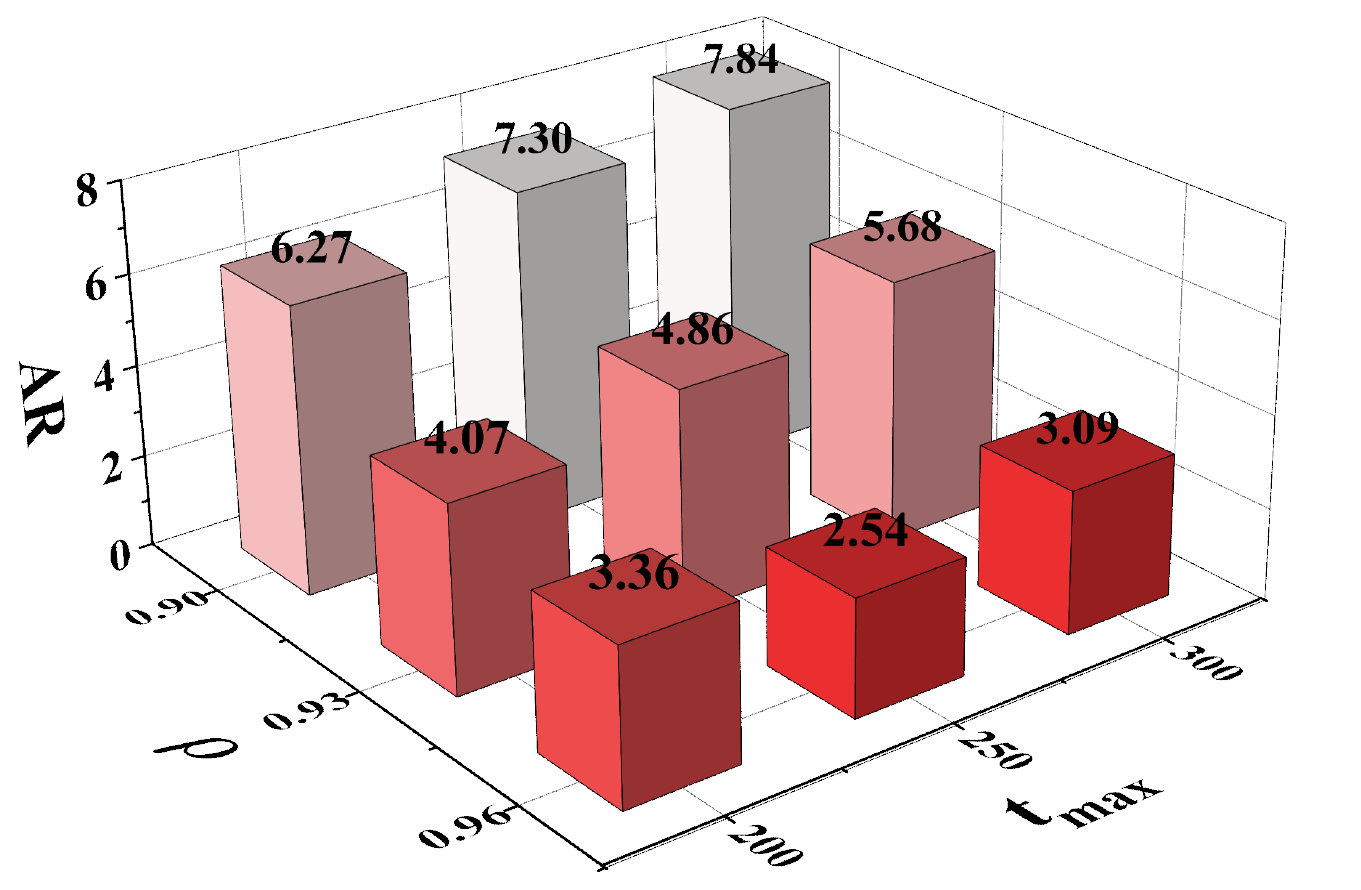} 
\caption{The AR results of GBO are based on 9 different parameter combinations.}
    \label{fig:柱状图}
\end{figure}

\noindent \textbf{Hyper-Parameter Sensitivity Analysis. }We study the effect of different parameter combinations on the performance of GBO. Specifically, we conducted 9 experiments with $\left\{ \rho \text{, }t_{\max} \right\} \in \left\{ 0.90\text{, 0.93, }0.96 \right\} \times \left\{ 200\text{, 250, }300 \right\}$.
The AR for the combination of these 9 parameters is depicted in Figure \ref{fig:柱状图}.

From our observation, GBO has the best performance when $\rho$ and $t_{max}$ are equal to 0.96 and 250, respectively. Under this combination of parameters, when GBO converges, the radius of the granular-ball becomes $10^{-5}$ of the initial radius. However, when $\rho$ and $t_{max}$ are equal to 0.90 and 300, respectively, the performance of GBO is the worst. Under this combination of parameters, when GBO converges, the radius of the granular-ball becomes $10^{-14}$ of the initial radius. When $\rho$ is smaller, the larger $t_{max}$ is, the less effective GBO is. However, performance does not always improve when $\rho$ is larger, and when $t_{max}$ is also increased. This phenomenon may be attributed to the fact that when the algorithm converges, the radius of the granular-ball should be within a suitable range, otherwise if the radius of the granular-ball is too small, then on the one hand, there is not much need to consume the number of fitness evaluations. On the other hand, approaching the local minimum too precisely may cause the algorithm to fall into the local minimum, which will negatively affect the optimization performance of the algorithm.

\begin{table}[!t]
    \centering
    \caption{Detailed comparison between SAMODE, GA-MPC, and GBO on the Spread Spectrum Radar Polly Phase Code Design problem.}
    \begin{tabular}{lccc c}
        \toprule
        $fes$  & Metric & SAMODE & GA-MPC & GBO \\ 
        \midrule
        \multirow{5}{*}{50000} & Best   & 8.21E-01 & 7.75E-01 & \textbf{6.39E-01} \\
                               & Median & 1.27E+00 & 1.74E+00 & \textbf{7.92E-01} \\
                               & Worst  & 1.70E+00 & 1.92E+00 & \textbf{1.06E+00} \\
                               & Mean   & 1.29E+00 & 1.62E+00 & \textbf{8.07E-01} \\
                               & Std.   & 1.93E-01 & 3.24E-01 & 8.81E-03 \\
        \midrule
        \multirow{5}{*}{100000} & Best   & 5.08E-01 & 5.08E-01 & \textbf{5.00E-01} \\
                                & Median & 9.99E-01 & 7.95E-01 & \textbf{5.58E-01} \\
                                & Worst  & 1.33E+00 & 1.68E+00 & \textbf{7.69E-01} \\
                                & Mean   & 9.73E-01 & 8.58E-01 & \textbf{5.80E-01} \\
                                & Std.   & 1.79E-01 & 2.73E-01 & 6.54E-03 \\
        \midrule
        \multirow{5}{*}{150000} & Best   & \textbf{5.00E-01} & \textbf{5.00E-01} & \textbf{5.00E-01} \\
                                & Median & 8.40E-01 & 7.58E-01 & \textbf{5.47E-01} \\
                                & Worst  & 9.94E-01 & 9.33E-01 & \textbf{7.57E-01} \\
                                & Mean   & 8.17E-01 & 7.48E-01 &\textbf{ 5.73E-01} \\
                                & Std.   & 1.19E-01 & 1.25E-01 & 6.10E-03 \\
        \bottomrule
    \end{tabular}
    \label{case}
\end{table}

\noindent \textbf{Case Study.} We apply GBO to solve the Spread Spectrum Radar Polly Phase Code Design problem (Appendix 6.1). When designing a radar-system that uses pulse compression, great attention must be given to the choice of the appropriate waveform. Many methods of radar pulse modulation that make pulse compression possible are known. Polyphase codes are attractive as they have other lower side lobes in the compressed signal and easier implementation of digital processing techniques. Later Dukic and Do-brosavljevic \cite{dukic1990method} proposed a new method for polyphase pulse compression code synthesis, which is based on the properties of the aperiodic autocorrelation function and the assumption of coherent radar pulse processing in the =receiver. The problem optimization problem in continuous variables and with numerous local optima. 

We compared the results with the algorithms that performed well on this problem, namely GA-CMP \cite{elsayed2011ga} and SAMODE \cite{elsayed2011differential}. The parameters of GBO are $N=10$, $\rho=0.94$ (applicable for fes), the solution is performed with $t_{max}$ of 150, and $fes_{max}$ of each algorithm is 150,000. The results are shown in Table \ref{case} , from which it can be seen that GBO has shown great advantages in this practical problem.

\section{Conclusion}
This paper proposes a multi-granularity optimization algorithm (GBO) using granular-balls. GBO employs a splitting mechanism to cover the solution space, performing a global search from coarse to fine granularity, and finds the optimal solution through synergistic searches between granular-balls. This approach replaces traditional point-based and regional search methods, enabling a more comprehensive exploration of the solution space. Experiments on the CEC2013 benchmark and the Spread Spectrum Radar Polly Phase Code Design problem demonstrate GBO's superiority. However, we aim to develop a more adaptive method for adjusting the granular-ball radius to improve efficiency, which will be addressed in future work. 


\bibliographystyle{named}
\bibliography{ijcai25}

\section{Appendix}
\subsection{Problem Models}
The Spread Spectrum Radar Polly Phase Code Design problem can be expressed as follows: 

\begin{equation}
        \text{global} \ \underset{x \in X}{\min} f(x) = \max \{ \phi_1(x), \dots, \phi_{2m} \},
\end{equation}

\noindent where $X = \{ (x_1, \dots, x_n) \in \mathbb{R}^n \mid 0 \leq x_j \leq 2\pi, \, j = 1, \dots, n \},$ $m = 2n-1$ and $\phi$ satisfies the following formula:

\begin{equation}
    \phi_{2i-1}(x) = \sum_{j=i}^n \cos \left( \sum_{k=|2i-j-1|+1}^j x_k \right), 
    \label{eq1}
\end{equation}

\begin{equation}
    \phi_{2i}(x) = 0.5 + \sum_{j=i}^n \cos \left( \sum_{k=|2i-j|+1}^j x_k \right), 
    \label{eq2}
\end{equation}

\begin{equation}
    \phi_{m+i}(x) = -\phi_i(x),
    \label{eq3}
\end{equation}

\noindent where the index \(i\) in Equations (\ref{eq1}), (\ref{eq2}), and (\ref{eq3}) takes values in the ranges \(1 \leq i \leq n\), \(1 \leq i \leq n-1\), and \(1 \leq i \leq m\), respectively.

Here the objective is to minimize the module of the biggest among the samples of the so-called auto-correlation function which is related to the complex envelope of the compressed radar pulse at the optimal receiver output, while the variables represent symmetrized phase differences. This problem belongs to the class of continuous min–max global optimization problems. They are characterized by the fact that the objective function is piecewise smooth.

\begin{table*}[h]
        \setlength{\abovecaptionskip}{0.3cm}
        \setlength{\belowcaptionskip}{0.3cm}
        \scriptsize
        \centering
         \caption{Comparison of GBO with several classic optimization algorithms in $\textbf{30}$-$D$.}
        \setlength{\tabcolsep}{0.2mm}
            \fontsize{7}{10}\selectfont
            \resizebox{\textwidth}{!}{ 
        \begin{tabular}{|c|cc|cc|cc|cc|cc|cc|cc|}
            \hline
             \multirow{2}{*}{\textit{f}} & \multicolumn{2}{c|}{GBO}             & \multicolumn{2}{c|}{ABC}              & \multicolumn{2}{c|}{DE}             & \multicolumn{2}{c|}{GA}              & \multicolumn{2}{c|}{PSO}           & \multicolumn{2}{c|}{SHADE}           & \multicolumn{2}{c|}{LoTFWA}  \\ \cline{2-15} 
            & \multicolumn{1}{c|}{Mean} & Std.     & \multicolumn{1}{c|}{Mean} & Std.     & \multicolumn{1}{c|}{Mean} & Std.     & \multicolumn{1}{c|}{Mean} & Std.     & \multicolumn{1}{c|}{Mean} & Std.     & \multicolumn{1}{c|}{Mean} & Std.   & \multicolumn{1}{c|}{Mean} & Std.  \\ \hline
            $F_{1}$                          & 3.25E-05	& 5.79E-06
	& 4.55E-13- 	& 7.80E-14 		& \textbf{0.00E+00}- 	& \textbf{0.00E+00} 		&1.84E+00+	&5.09E-01		&2.77E+02+	&5.05E+02		& \textbf{0.00E+00}- 	& \textbf{0.00E+00} 		& 1.27E-12- 	& 9.85E-13  \\ 
				$F_{2} $                          & 8.30E+05	&4.46E+05
	&1.00E+07+	&2.78E+06		& 3.88E+05- 	& 2.31E+05 		&2.29E+07+	&1.08E+07		&4.70E+06+	&4.93E+06		& \textbf{1.26E+04}- 	& \textbf{1.05E+04} 		&9.55E+05$\approx$	&4.25E+05\\
				$F_{3}$                          & 3.37E+01	&4.37E+00
	&7.04E+08+	&4.86E+08		& \textbf{3.00E+01}- 	& \textbf{1.34E+02} 		&5.62E+08+	&5.26E+08		&1.11E+10+	&8.57E+09		&2.53E+05+	&1.26E+06		&3.22E+07+	&3.33E+07 \\
				$F_{4}$                          & 3.60E+04	&9.53E+03
	&7.58E+04+	&9.19E+03		& 1.54E+03- 	& 5.83E+02 		& 1.40E+04- 	& 4.16E+03 		& 3.36E+03- 	& 1.55E+03 		& \textbf{1.25E-04}- 	& \textbf{3.21E-04} 		& 1.97E+03- 	& 7.33E+02  \\
				$F_{5}$                          & 3.53E-03 & 4.12E-04
	& 7.80E-13- 	& 9.29E-14 		& \textbf{9.14E-14-} 	& \textbf{4.51E-14} 		&1.20E+00+	&2.26E-01		&4.19E+02+	&6.24E+02		& 1.14E-13- 	& 1.26E-29 		&4.09E-03+	&6.43E-04 \\
				$F_{6}$                          & 2.08E+01	&1.89E+01
	&1.41E+01$\approx$	&4.59E+00		& 1.03E+01- 	& 4.79E+00 		&6.64E+01+	&2.68E+01		&7.24E+01+	&4.00E+01		& \textbf{5.18E-01}- 	& \textbf{3.66E+00} 		&1.56E+01$\approx$	&9.98E+00 \\
				$F_{7}$                          & \textbf{6.29E-02}	&\textbf{8.95E-02}
	&1.17E+02+	&1.47E+01		&3.73E-01+	&7.13E-01		&5.37E+01+	&1.28E+01		&1.83E+02+	&1.08E+02		&4.52E+00+	&5.20E+00		&5.36E+01+	&1.37E+01\\
				$F_{8}$                          & 2.09E+01	&5.81E-02
	&2.09E+01$\approx$	&4.59E-02		&2.09E+01$\approx$	&4.49E-02		&2.10E+01+	&4.80E-02		&2.09E+01$\approx$	&6.19E-02		& \textbf{2.08E+01}- 	& \textbf{1.64E-01} 		&2.09E+01$\approx$	&6.61E-02 \\
				$F_{9}$                          & \textbf{3.66E+00}	&\textbf{1.93E+00}
	&2.99E+01+	&1.63E+00		&3.76E+01+	&4.48E+00		&2.37E+01+	&2.42E+00		&3.48E+01+	&3.03E+00		&2.78E+01+	&1.60E+00		&1.71E+01+	&2.07E+00\\
				$F_{10}$                         & \textbf{1.66E-03}	&\textbf{2.72E-03}
	&1.84E+00+	&4.33E-01		&7.34E-03+	&7.73E-03		&3.08E+01+	&1.37E+01		&1.56E+02+	&1.30E+02		&6.81E-02+	&3.18E-02		&2.93E-02+	&1.53E-02	\\
				$F_{11}$                         & 2.10E+01	&4.96E+00
	& 1.10E-13- 	& 2.08E-14 		&1.24E+02+	&2.93E+01		& 1.78E+00- 	& 5.05E-01 		&2.67E+02+	&5.85E+01		& \textbf{0.00E+00}- 	& \textbf{0.00E+00} 		&8.78E+01+	&1.46E+01	\\
				$F_{12}$                         & \textbf{2.13E+01}	&\textbf{4.99E+00}
	&2.73E+02+	&3.97E+01		&1.81E+02+	&9.94E+00		&7.90E+01+	&1.79E+01		&3.06E+02+	&7.95E+01		&2.26E+01$\approx$	&3.85E+00		&8.68E+01+	&1.67E+01\\
				$F_{13}$                         & \textbf{3.85E+01}	&\textbf{1.25E+01}
	&3.10E+02+	&3.02E+01		&1.79E+02+	&9.32E+00		&1.57E+02+	&3.11E+01		&3.82E+02+	&6.88E+01		&4.99E+01+	&1.27E+01		&1.64E+02+	&1.75E+01\\
				$F_{14}$                         & 7.73E+02	&2.74E+02
	& 2.37E+00- 	& 1.46E+00 		&5.38E+03+	&5.41E+02		& 1.12E+01- 	& 2.82E+00 		&3.98E+03+	&8.49E+02		& \textbf{3.88E-02}- 	& \textbf{2.40E-02} 		&2.78E+03+	&2.80E+02\\
				$F_{15}$                         & \textbf{7.88E+02}	&\textbf{2.56E+02}
	&3.85E+03+	&2.98E+02		&7.13E+03+	&2.64E+02		&4.25E+03+	&6.34E+02		&4.50E+03+	&6.37E+02		&3.36E+03+	&3.12E+02		&2.77E+03+	&2.59E+02\\
				$F_{16}$                         & \textbf{6.71E-03}	&\textbf{2.57E-03}
	&1.39E+00+	&2.05E-01		&2.48E+00+	&2.79E-01	 	&1.67E+00+	&3.96E-01	 	&1.48E+00+	&3.76E-01	 	&1.00E+00+	&1.89E-01	 	&1.59E-01+	&5.19E-02 \\
				$F_{17}$                         & 4.84E+01	&3.93E+00
	& 3.05E+01- 	& 4.14E-02 	 	&1.85E+02+	&1.56E+01	 	& 3.65E+01- 	& 1.02E+00 		&3.94E+02+	&7.39E+01		& \textbf{3.04E+01}- 	& \textbf{1.38E-14} 		&1.34E+02+	&2.60E+01 \\
				$F_{18}$                         & \textbf{4.78E+01}	&\textbf{3.65E+00}
	&3.01E+02+	&3.05E+01		&2.11E+02+	&9.98E+00		&1.90E+02+	&2.24E+01		&4.10E+02+	&7.85E+01		&7.31E+01+	&4.80E+00		&1.44E+02+	&2.21E+01\\
				$F_{19}$                         & 3.00E+00	&5.43E-01
	& \textbf{4.50E-01}- 	& \textbf{1.18E-01} 		&1.50E+01+	&1.08E+00		& 2.00E+00- 	& 2.90E-01 		&6.33E+01+	&1.63E+02		& 1.36E+00- 	& 1.11E-01 		&4.81E+00+	&8.81E-01	\\
				$F_{20}$                         & \textbf{8.68E+00}	&\textbf{6.81E-01}
	&1.44E+01+	&2.86E-01	 	&1.23E+01+	&2.69E-01	 	&1.19E+01+	&4.52E-01	 	&1.41E+01+	&5.72E-01	 	&1.10E+01+	&4.79E-01	 	&1.30E+01+	&1.14E+00	\\
				$F_{21}$                         & 2.48E+02	&9.67E+01
	& \textbf{1.78E+02}- 	& \textbf{3.16E+01 	}	&2.77E+02$\approx$	&6.18E+01		&3.24E+02+	&6.79E+01		&3.50E+02+	&1.10E+02		&2.96E+02+	&5.63E+01		& 2.02E+02- 	& 4.18E+01 \\
				$F_{22}$                         & 8.62E+02	&2.21E+02
	& \textbf{3.50E+01}- 	& \textbf{1.84E+01} 		&5.24E+03+	&8.11E+02	 	& 1.29E+02- 	& 4.09E+01 	 	&4.59E+03+	&1.02E+03	 	& 8.50E+01- 	& 4.09E+01 	 	&3.31E+03+	&4.09E+02\\
				$F_{23}$                         & \textbf{8.57E+02}	&\textbf{2.66E+02}
	&4.80E+03+	&4.81E+02		&7.19E+03+	&2.54E+02	 	&4.44E+03+	&6.21E+02	 	&5.68E+03+	&8.87E+02	 	&3.61E+03+	&4.39E+02	 	&3.32E+03+	&4.02E+02\\
				$F_{24}$                         & \textbf{2.00E+02}	&\textbf{1.45E-02}
	&2.87E+02+	&1.00E+01		&2.25E+02+	&1.26E+01	 	&2.63E+02+	&1.12E+01	 	&3.11E+02+	&1.07E+01	 	&2.15E+02+	&1.38E+01	 	&2.42E+02+	&7.46E+00\\
				$F_{25}$                         & \textbf{2.12E+02}	&\textbf{1.94E+01}
	&3.06E+02+	&4.65E+00		&2.45E+02+	&5.76E+00	 	&2.80E+02+	&9.25E+00	 	&3.32E+02+	&1.45E+01	 	&2.79E+02+	&9.04E+00	 	&2.78E+02+	&9.95E+00\\
				$F_{26}$                         & \textbf{2.00E+02}	&\textbf{2.56E-02}
	&2.01E+02+	&2.01E-01		&2.03E+02+	&1.80E+01		&2.11E+02+	&3.87E+01		&3.17E+02+	&8.98E+01		&2.08E+02+	&3.30E+01		&\textbf{2.00E+02}$\approx$	&\textbf{2.06E-02 }\\
				$F_{27}$                         & \textbf{3.03E+02}	&\textbf{1.80E-01}
	&4.00E+02+	&4.22E-01		&5.87E+02+	&1.16E+02		&9.28E+02+	&7.09E+01		&1.25E+03+	&9.24E+01		&8.24E+02+	&1.49E+02		&7.80E+02+	&5.91E+01	\\
				$F_{28}$                         & 3.00E+02	&1.25E-02
	& \textbf{2.11E+02}- 	& \textbf{7.73E+01} 		&3.00E+02$\approx$	&5.68E-14		&3.57E+02+	&1.00E+01	&1.85E+03+	&1.14E+03		&3.00E+02$\approx$	&1.19E-13		& 2.49E+02- 	& 8.71E+01 	\\ \hline
				\textit{w/t/l}              & \multicolumn{2}{c|}{-}               & \multicolumn{2}{c|}{17/2/9}          & \multicolumn{2}{c|}{19/3/6}          & \multicolumn{2}{c|}{21/1/6}          & \multicolumn{2}{c|}{26/1/1}          & \multicolumn{2}{c|}{15/2/11}         & \multicolumn{2}{c|}{20/4/4} \\ \hline
				\textit{Rank}               & \multicolumn{2}{c|}{\textbf{2.52 }}            & \multicolumn{2}{c|}{4.04}            & \multicolumn{2}{c|}{4.16}        & \multicolumn{2}{c|}{4.75}            & \multicolumn{2}{c|}{6.43}            & \multicolumn{2}{c|}{2.52}            &\multicolumn{2}{c|}{3.59} \\ \hline
			\end{tabular}}
			\label{accuracy_30D}
\end{table*}

\subsection{Convergence Properties} 
\begin{figure*}[htbp]
    \centering
   \includegraphics[width=0.23\textwidth]{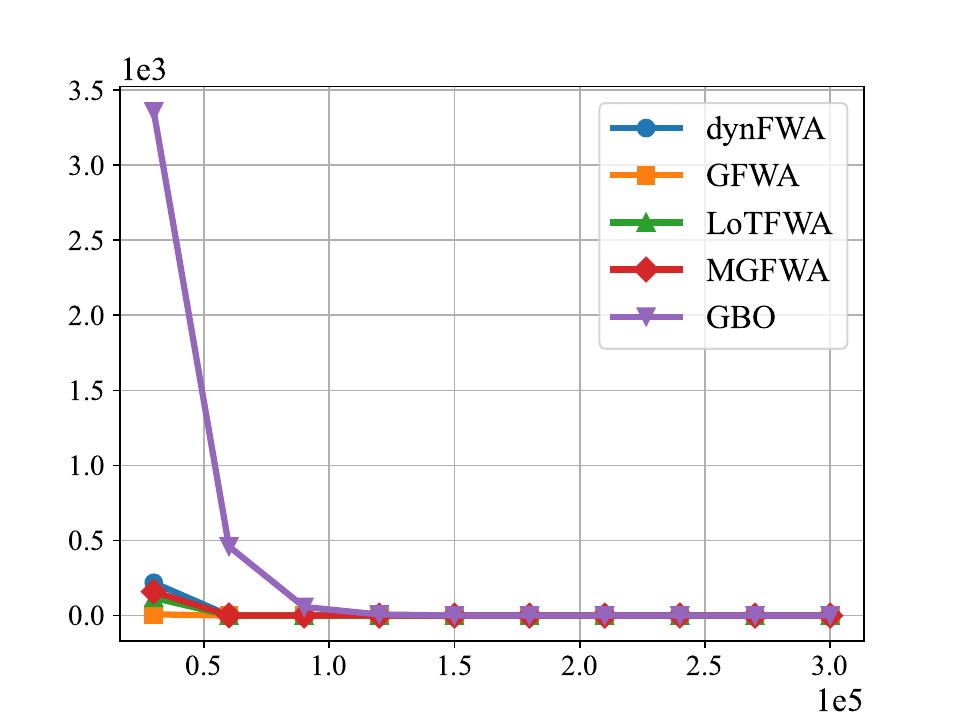}
\includegraphics[width=0.23\textwidth]{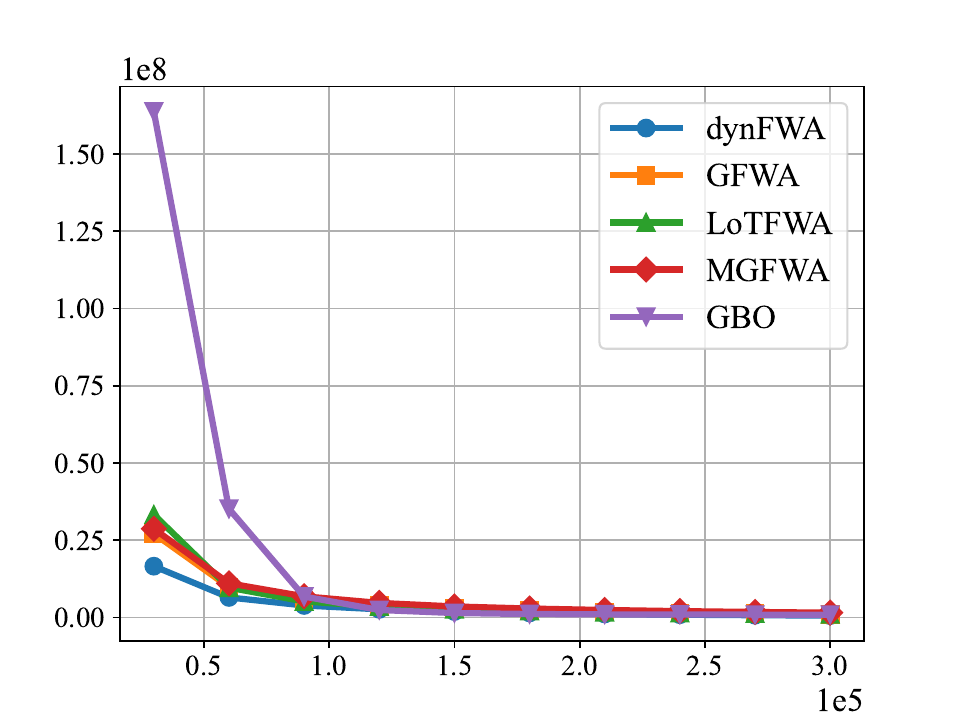}
\includegraphics[width=0.23\textwidth]{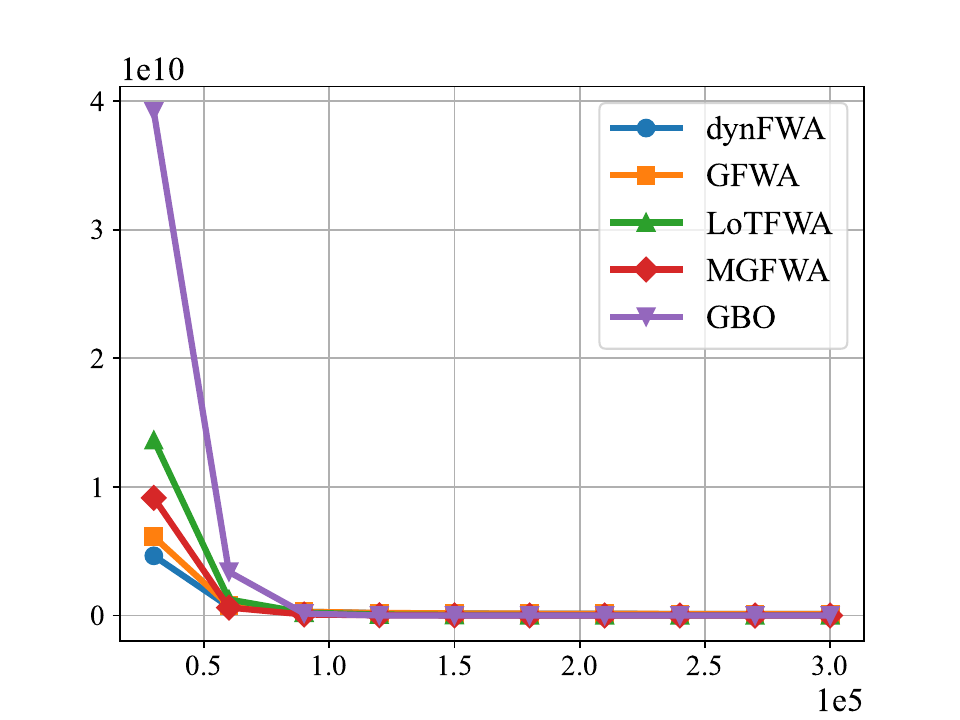}
\includegraphics[width=0.23\textwidth]{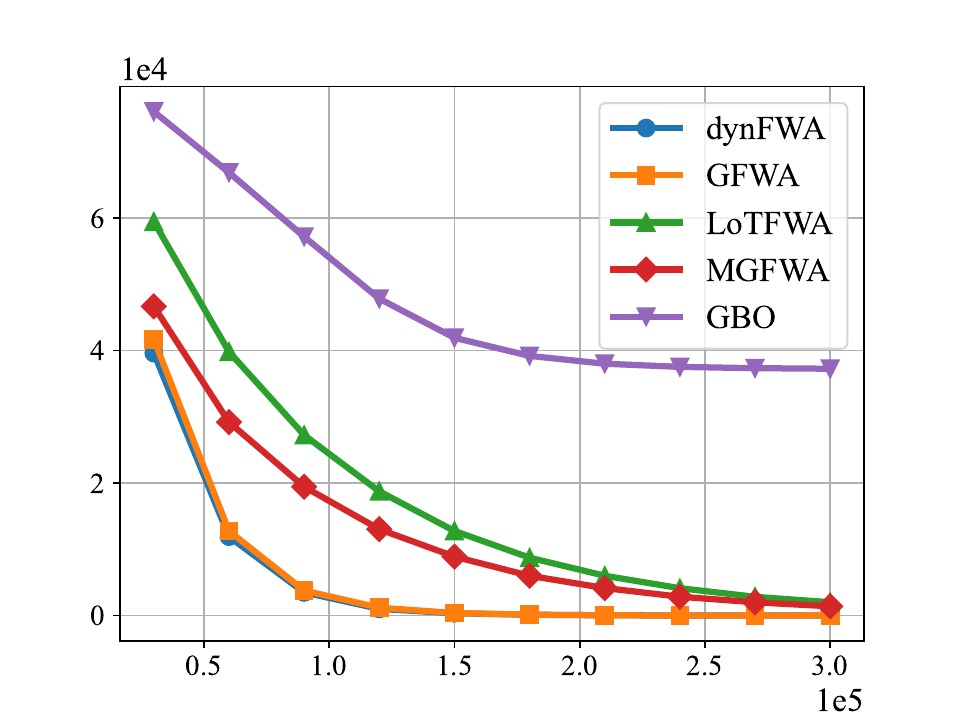}

\includegraphics[width=0.23\textwidth]{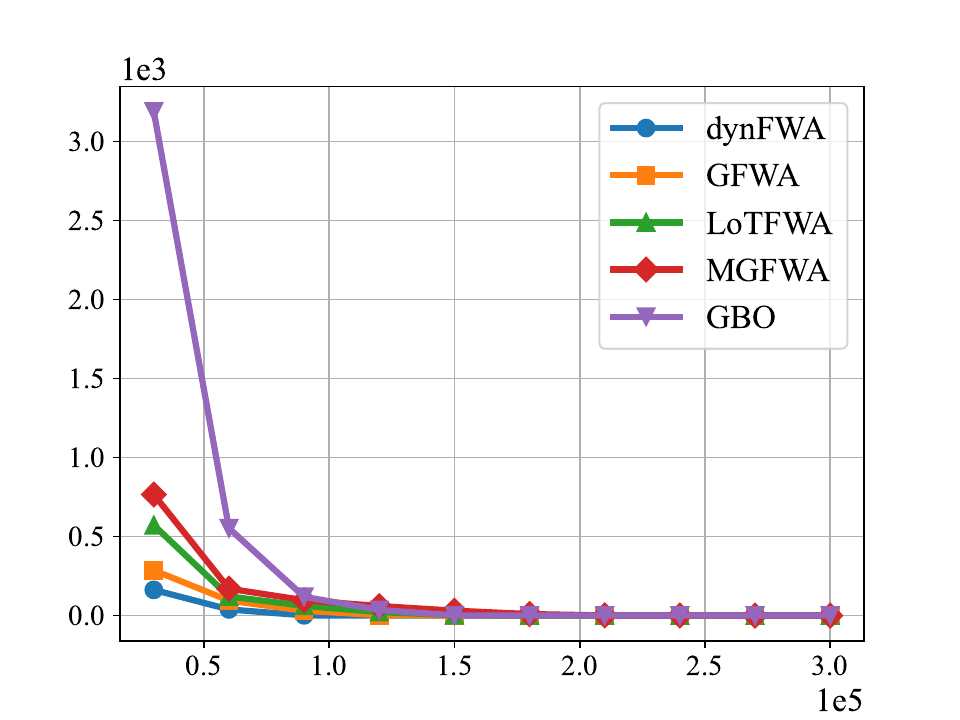}
\includegraphics[width=0.23\textwidth]{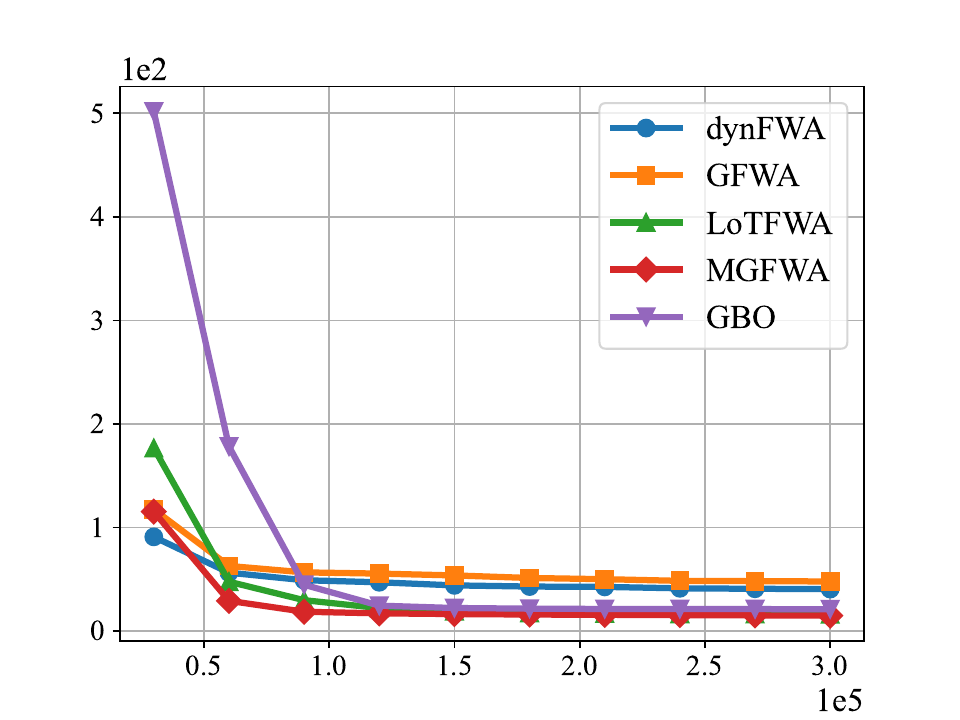}
\includegraphics[width=0.23\textwidth]{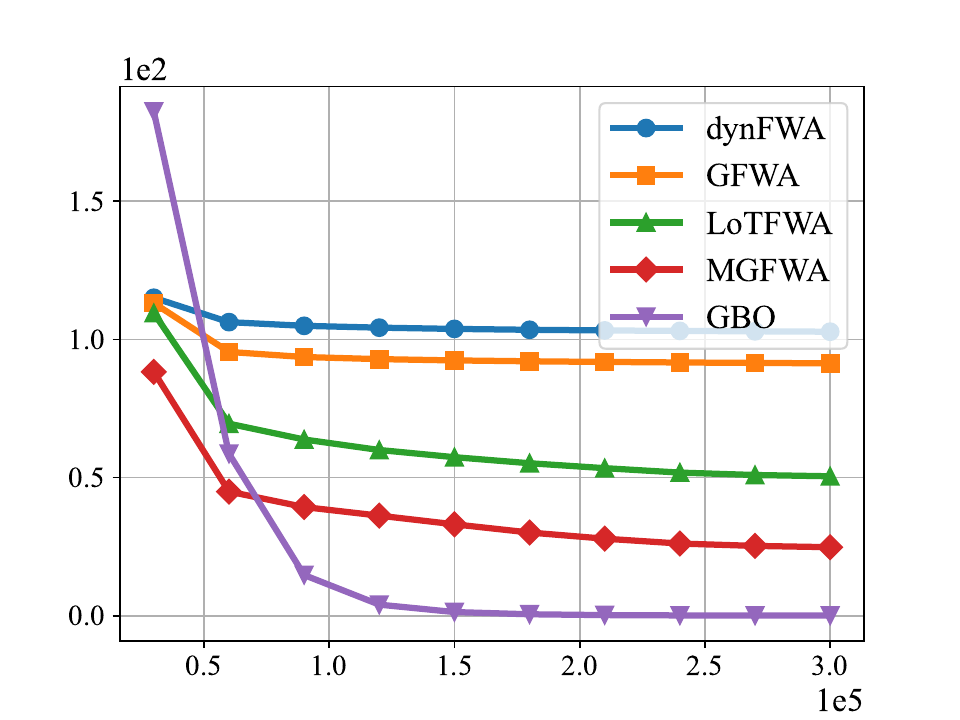}
\includegraphics[width=0.23\textwidth]{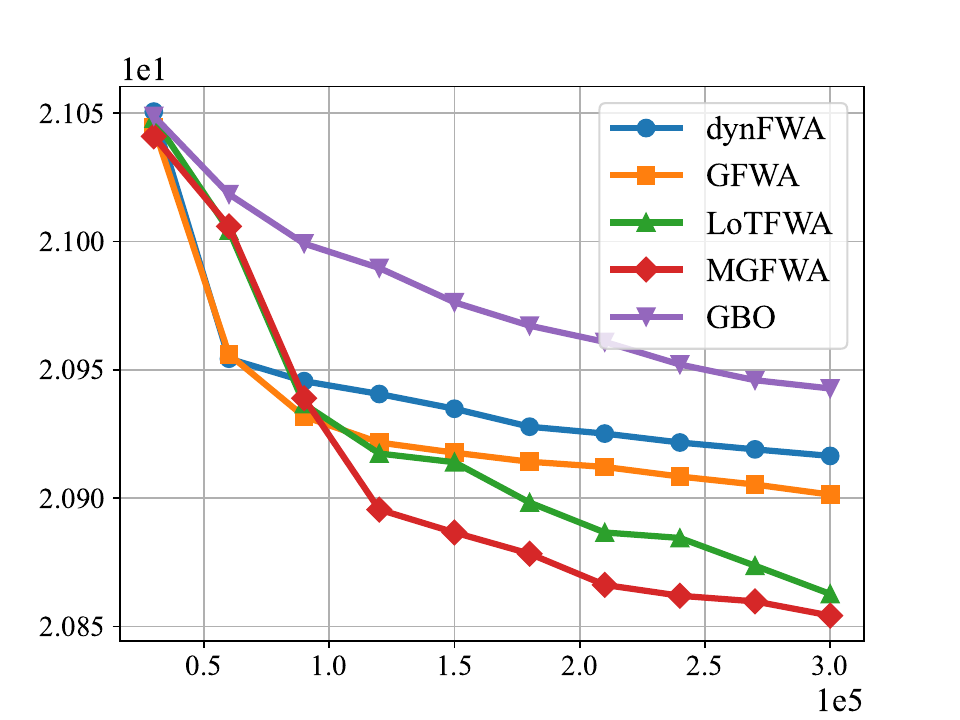}

\includegraphics[width=0.23\textwidth]{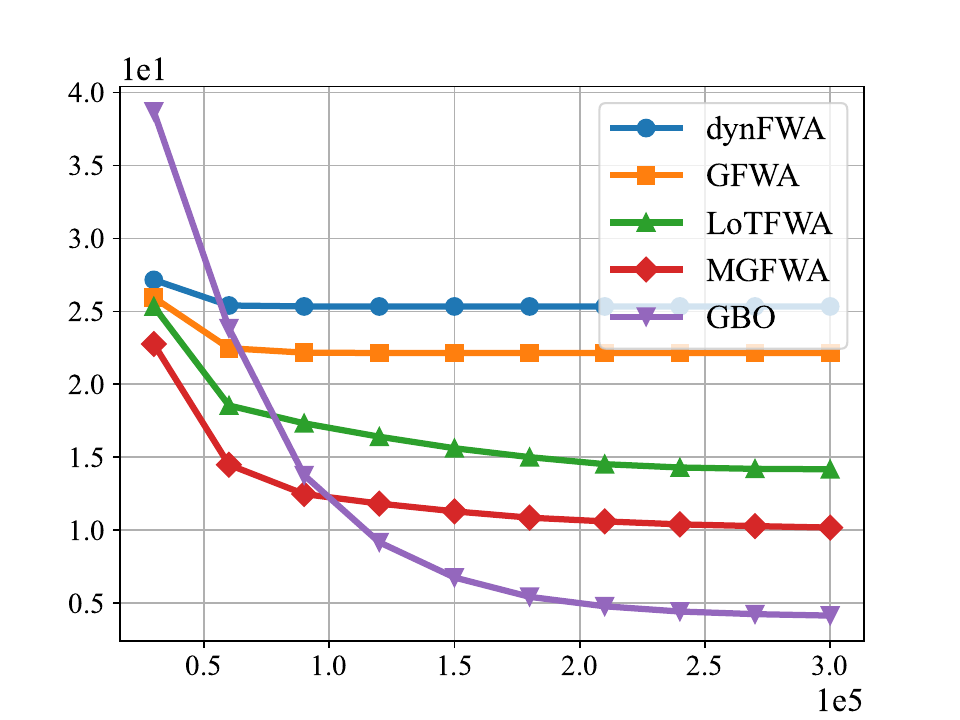}
\includegraphics[width=0.23\textwidth]{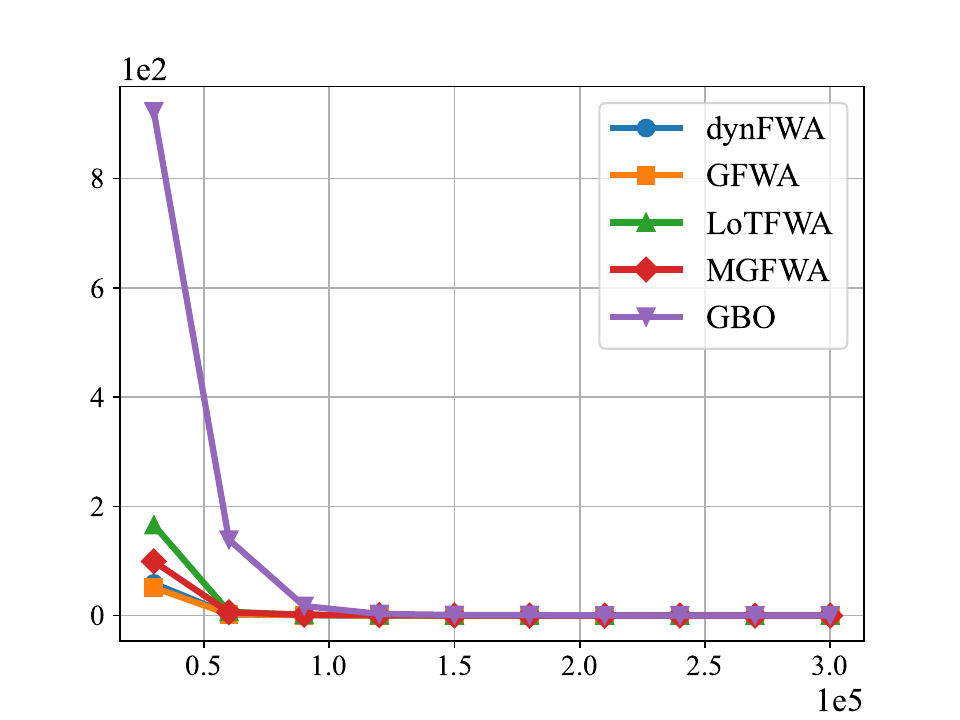}
\includegraphics[width=0.23\textwidth]{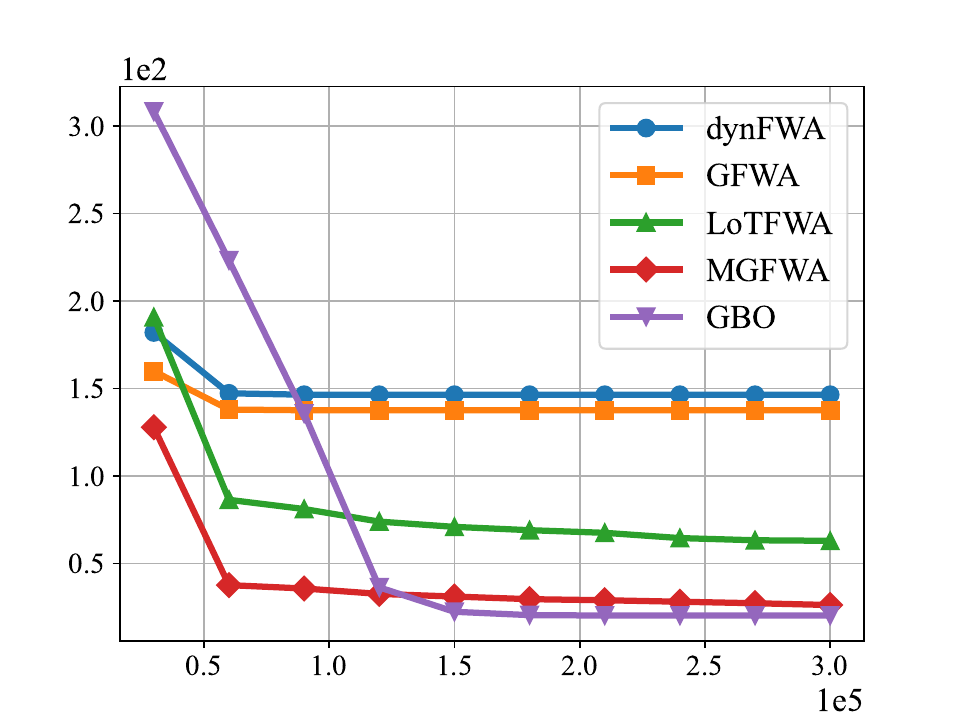}
\includegraphics[width=0.23\textwidth]{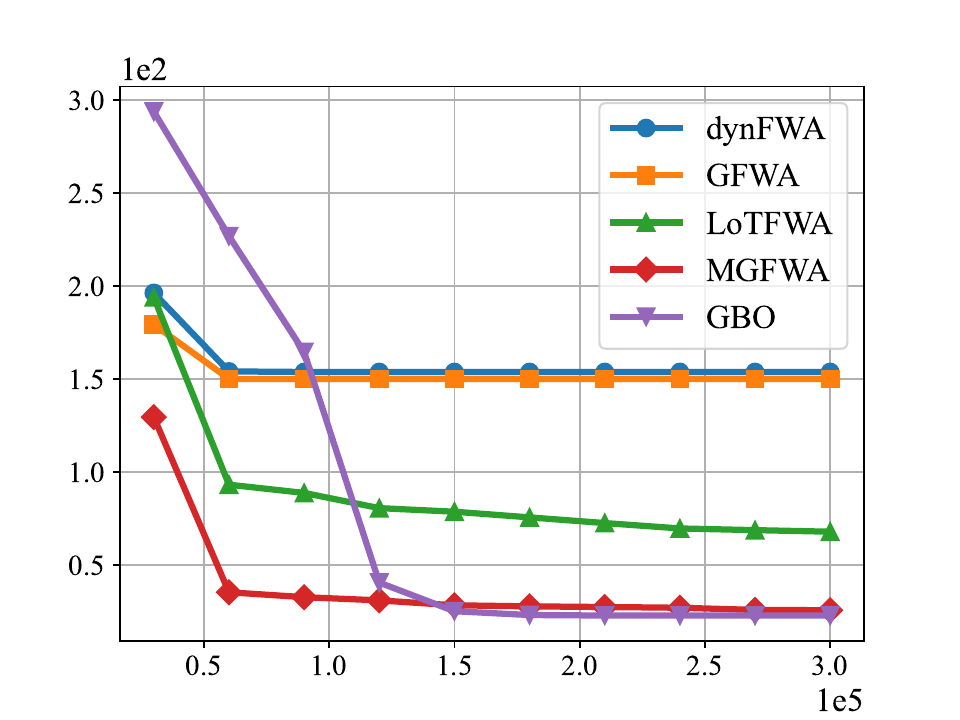}

\includegraphics[width=0.23\textwidth]{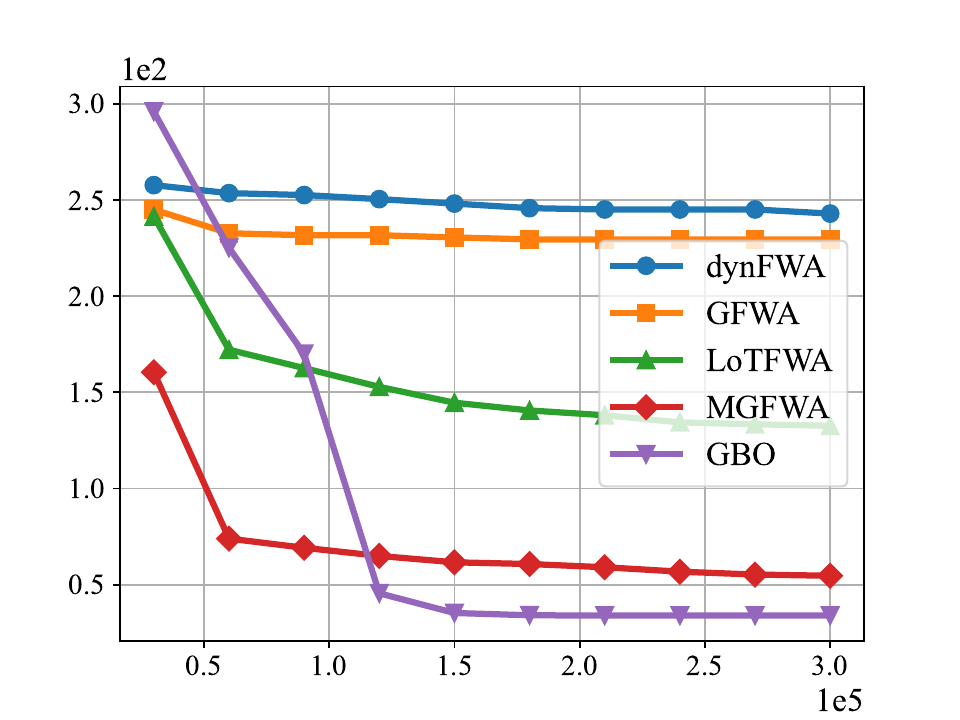}
\includegraphics[width=0.23\textwidth]{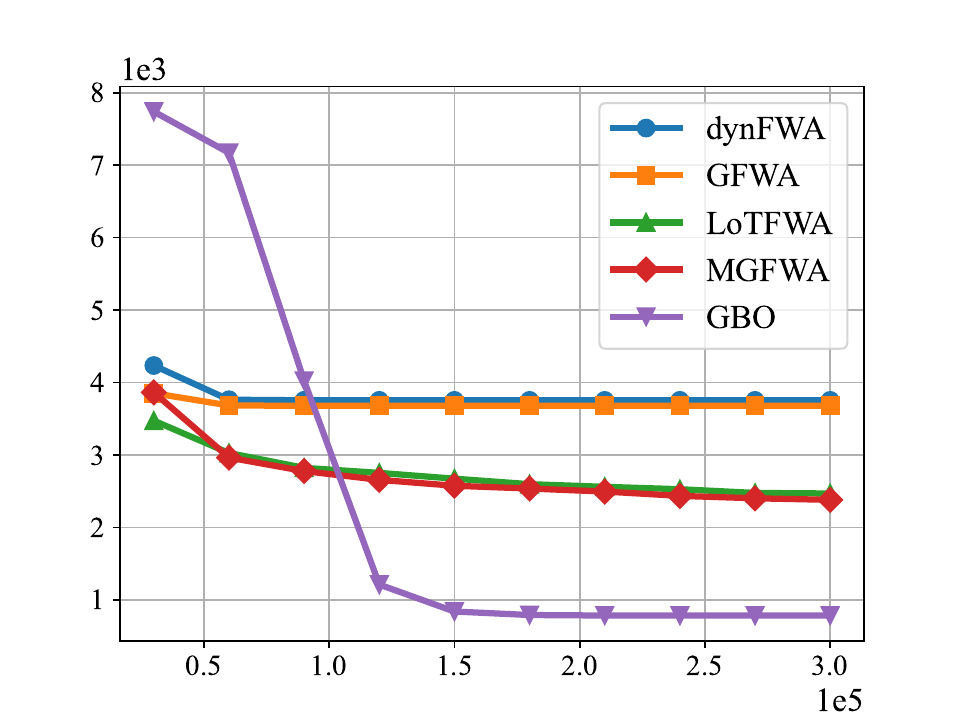}
\includegraphics[width=0.23\textwidth]{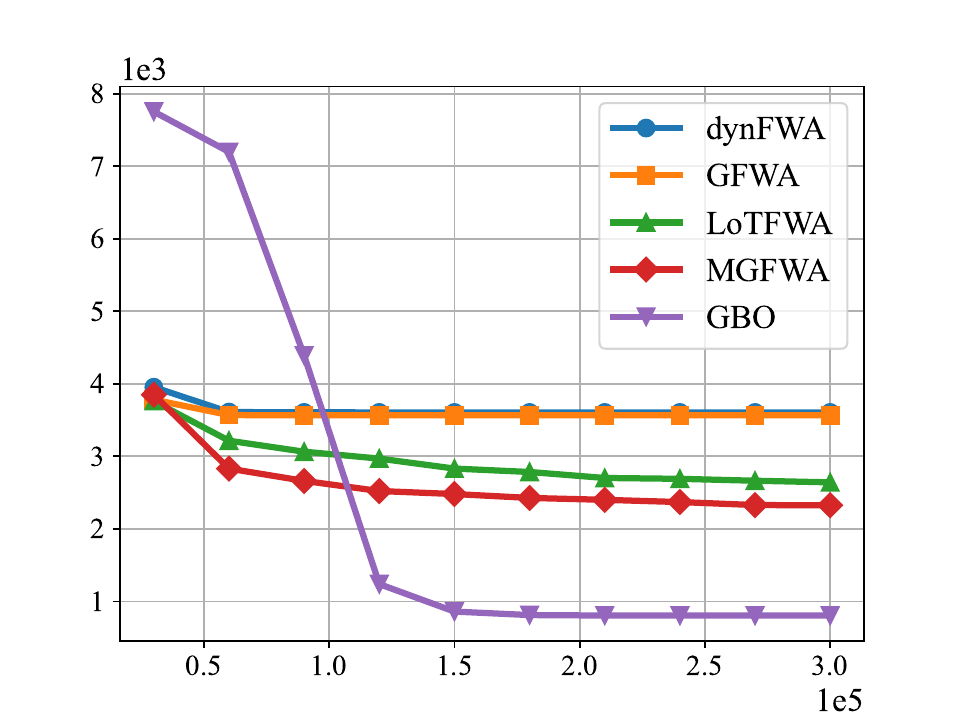}
\includegraphics[width=0.23\textwidth]{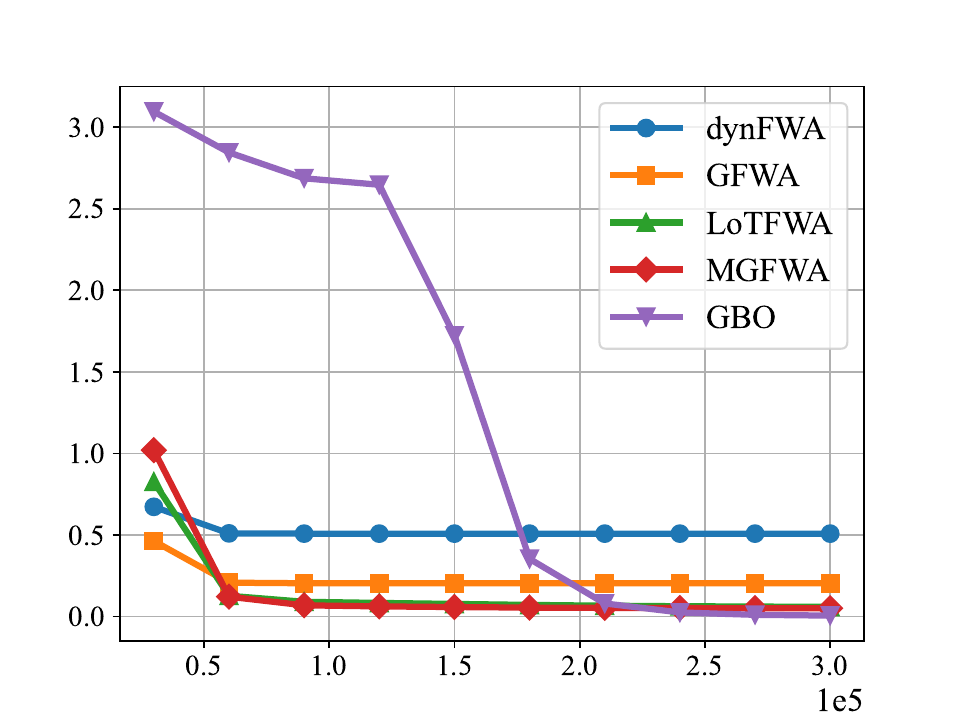}

\includegraphics[width=0.23\textwidth]{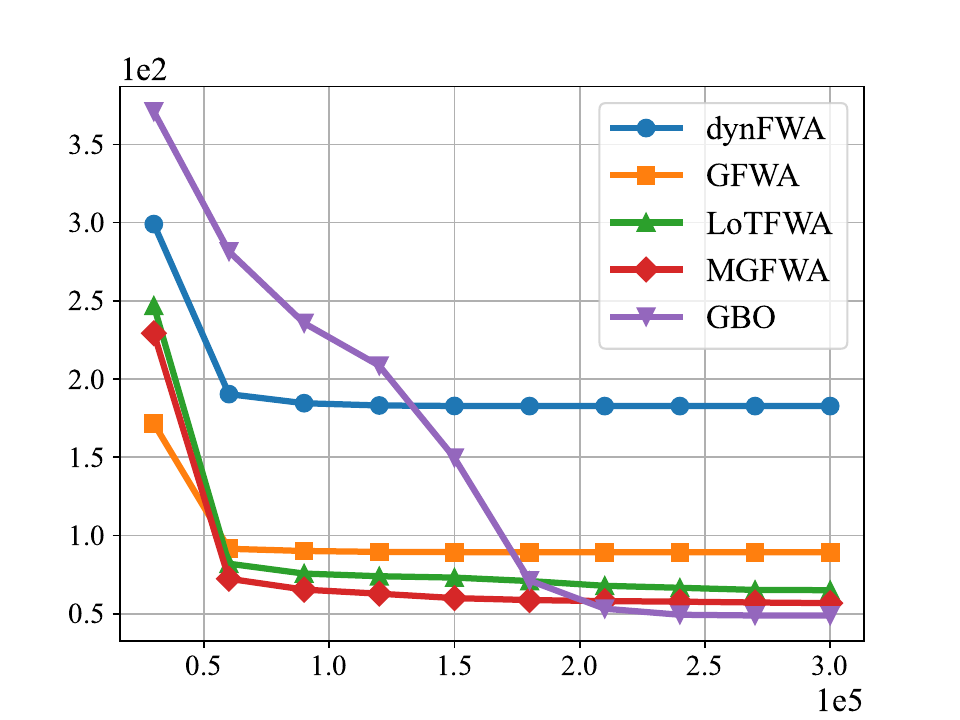}
\includegraphics[width=0.23\textwidth]{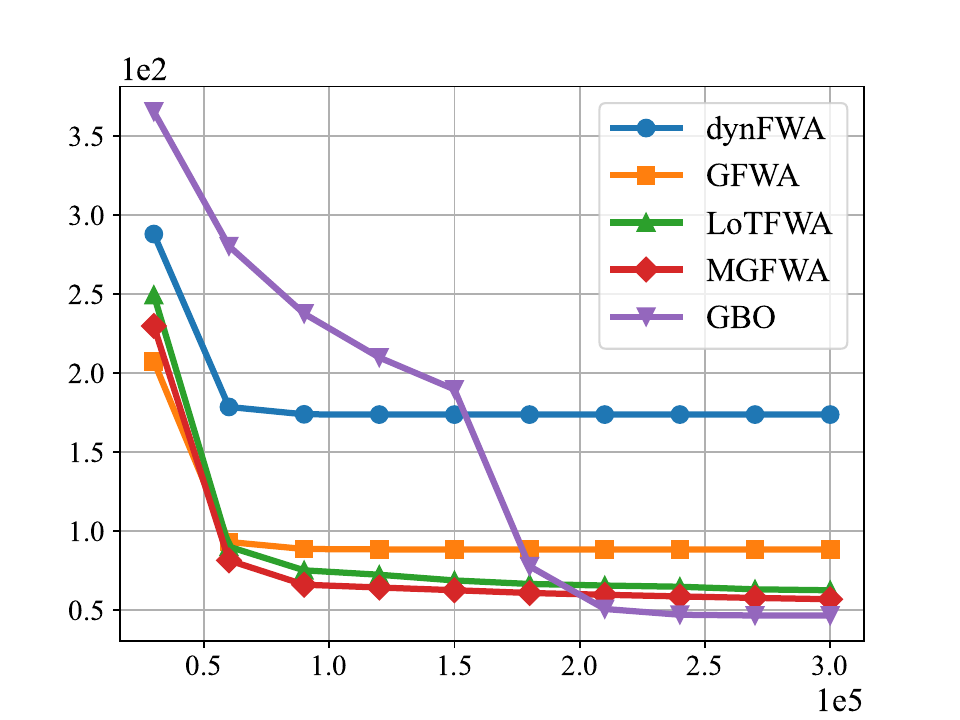}
\includegraphics[width=0.23\textwidth]{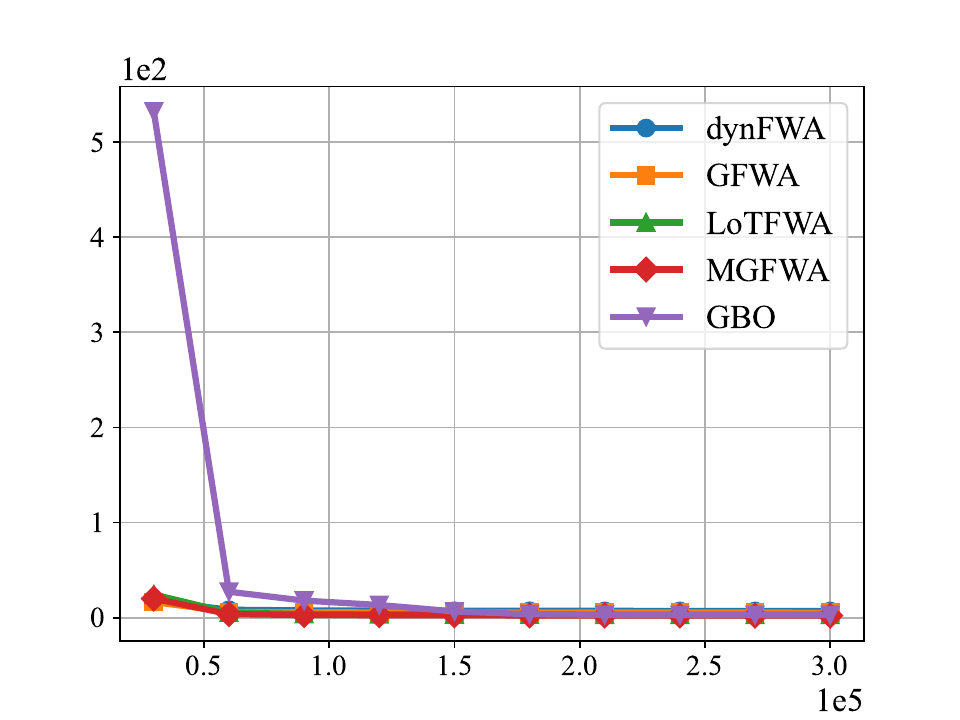}
\includegraphics[width=0.23\textwidth]{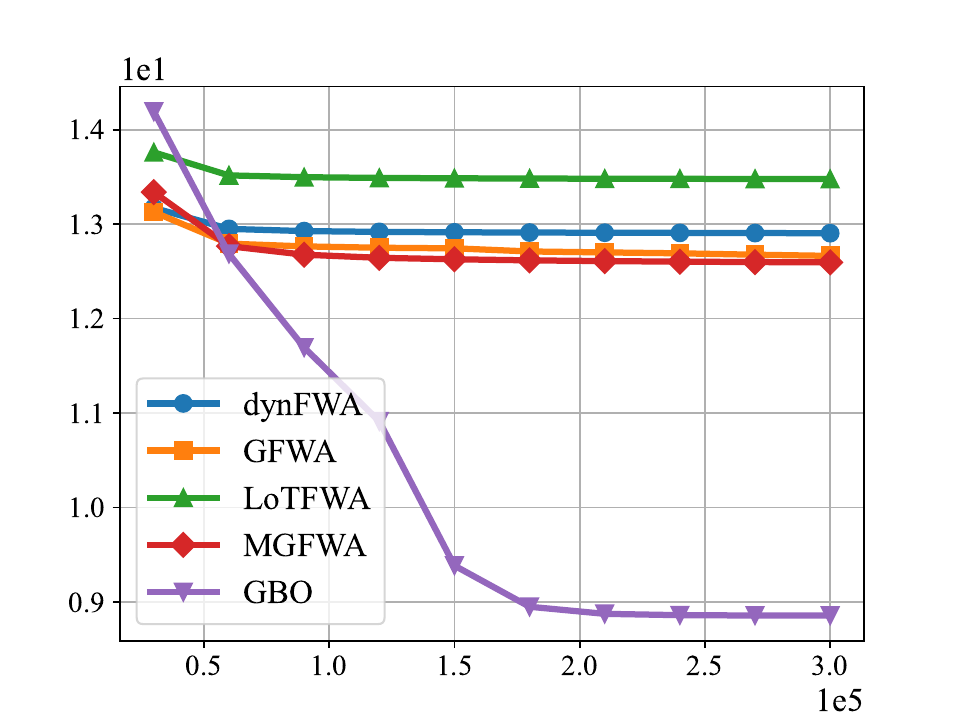}

\includegraphics[width=0.23\textwidth]{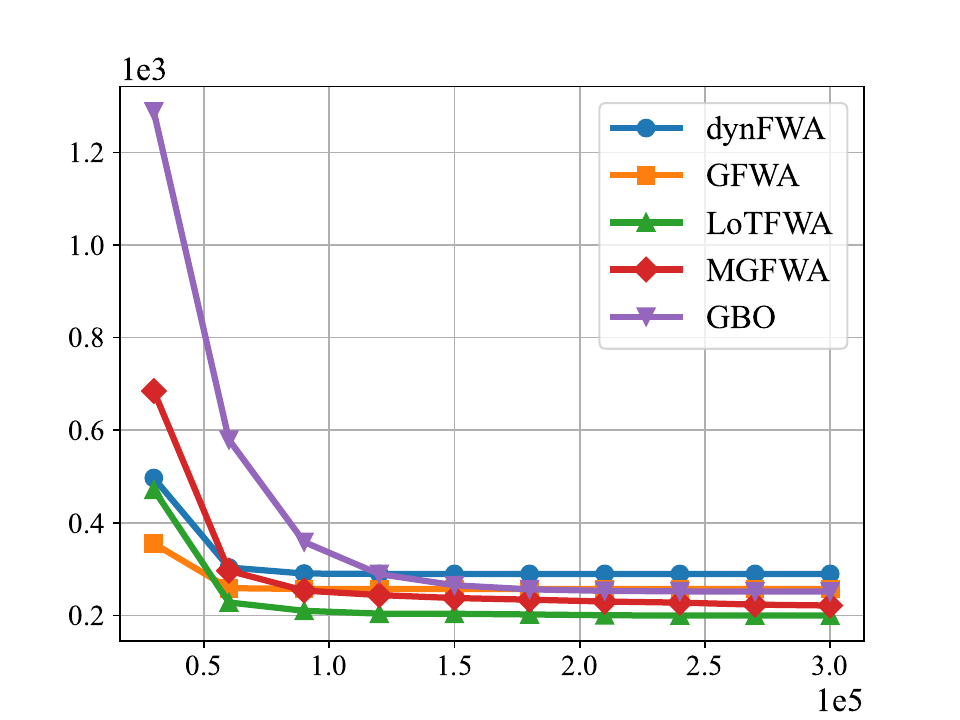}
\includegraphics[width=0.23\textwidth]{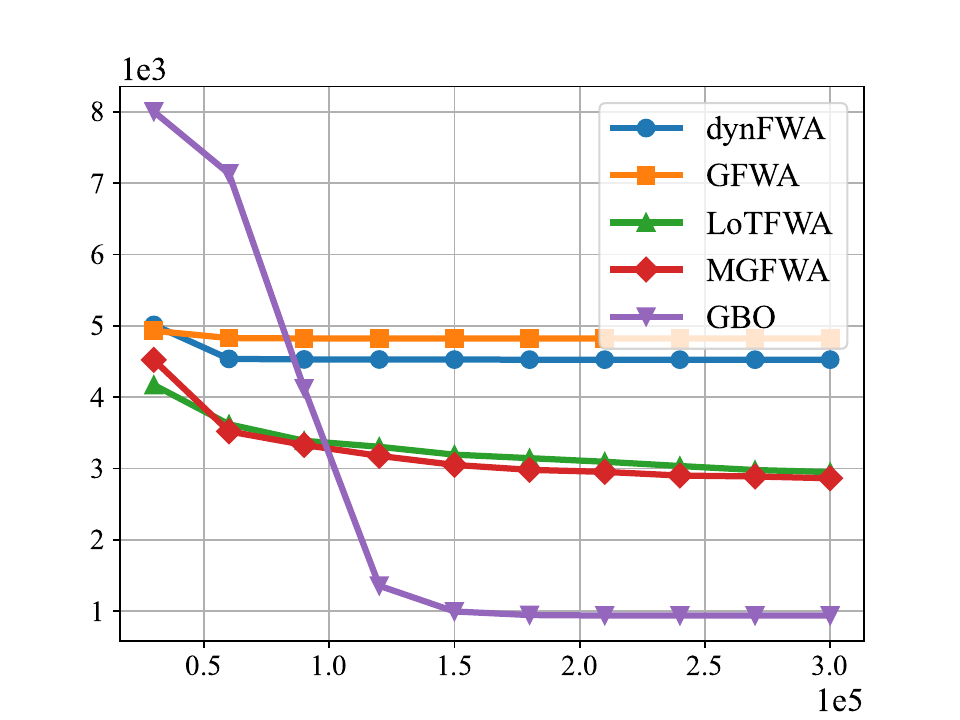}
\includegraphics[width=0.23\textwidth]{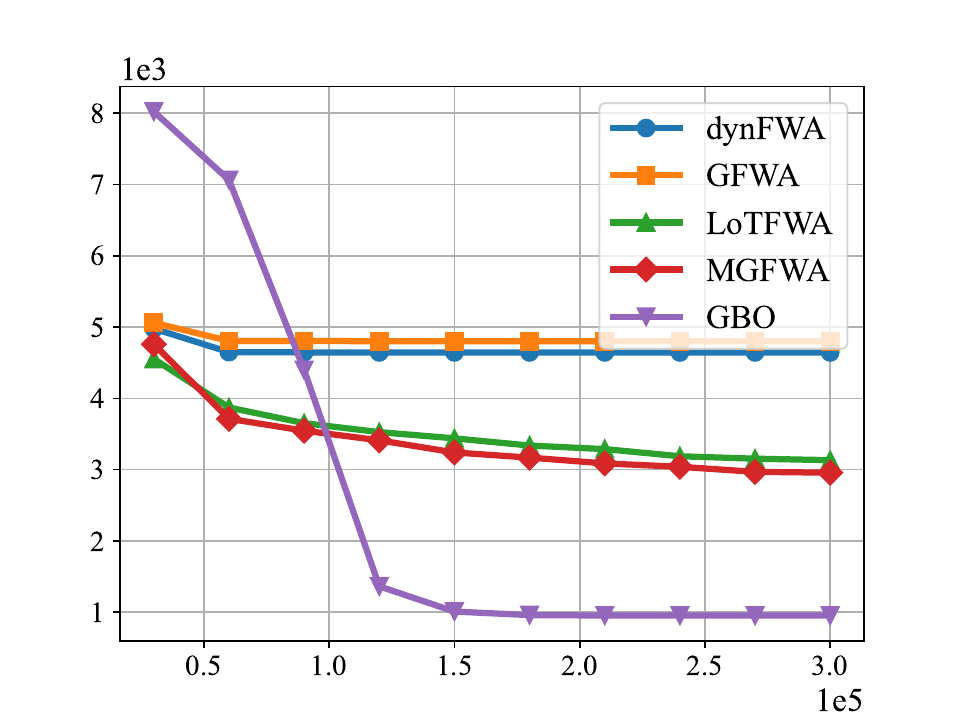}
\includegraphics[width=0.23\textwidth]{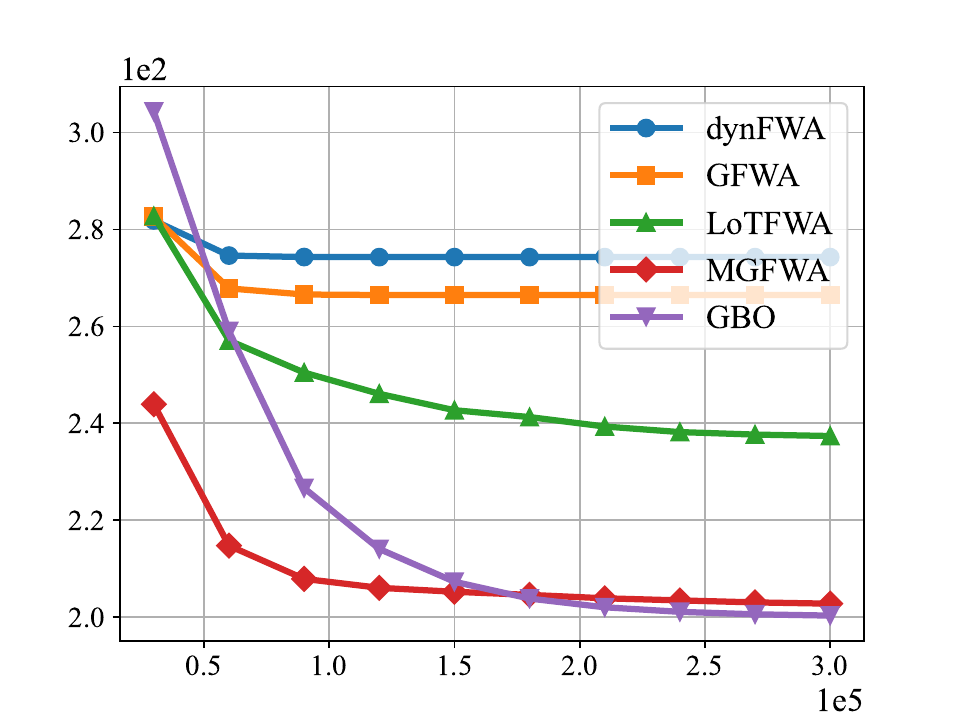}

\includegraphics[width=0.23\textwidth]{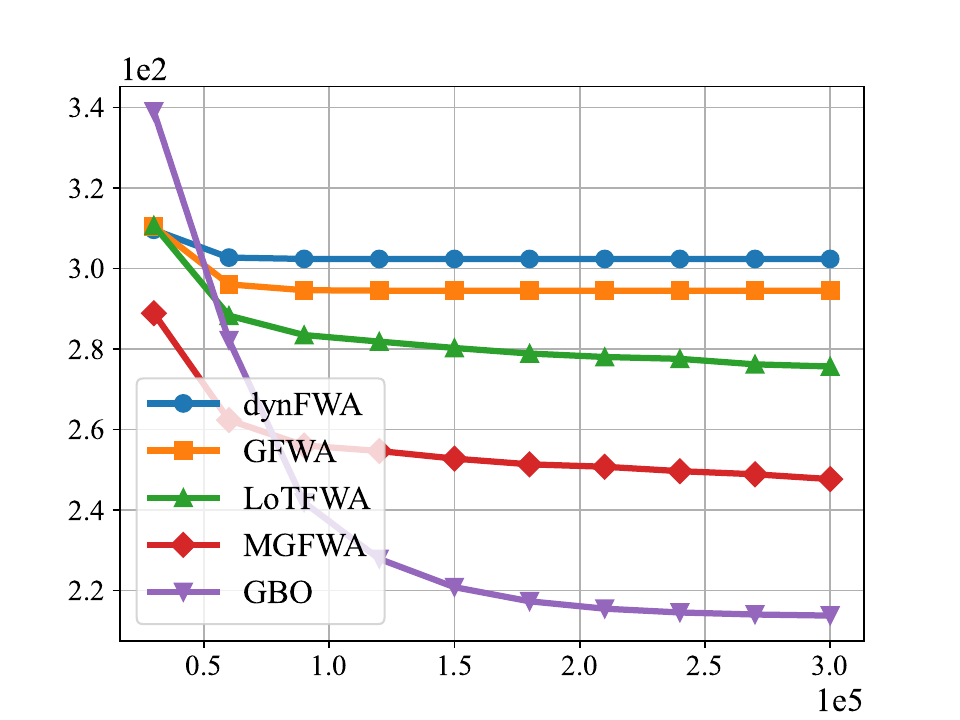}
\includegraphics[width=0.23\textwidth]{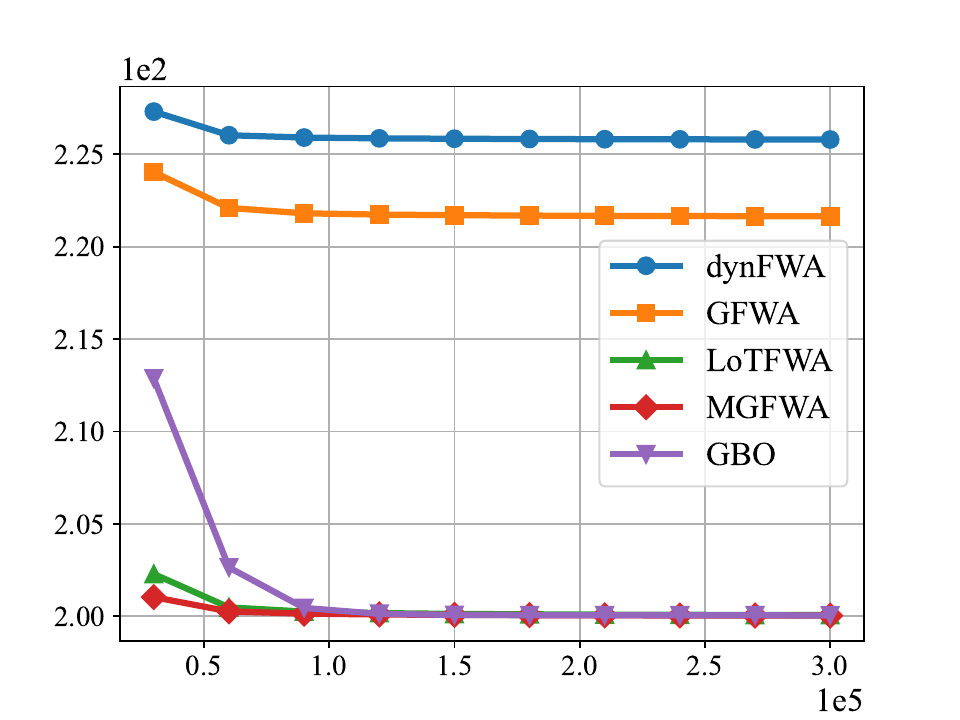}
\includegraphics[width=0.23\textwidth]{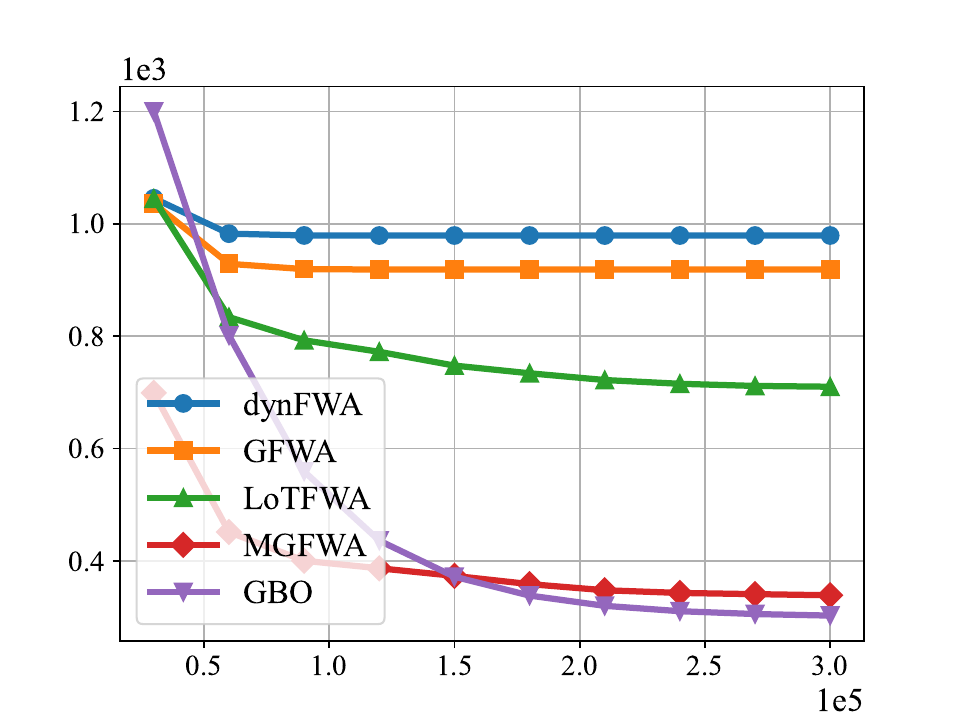}
\includegraphics[width=0.23\textwidth]{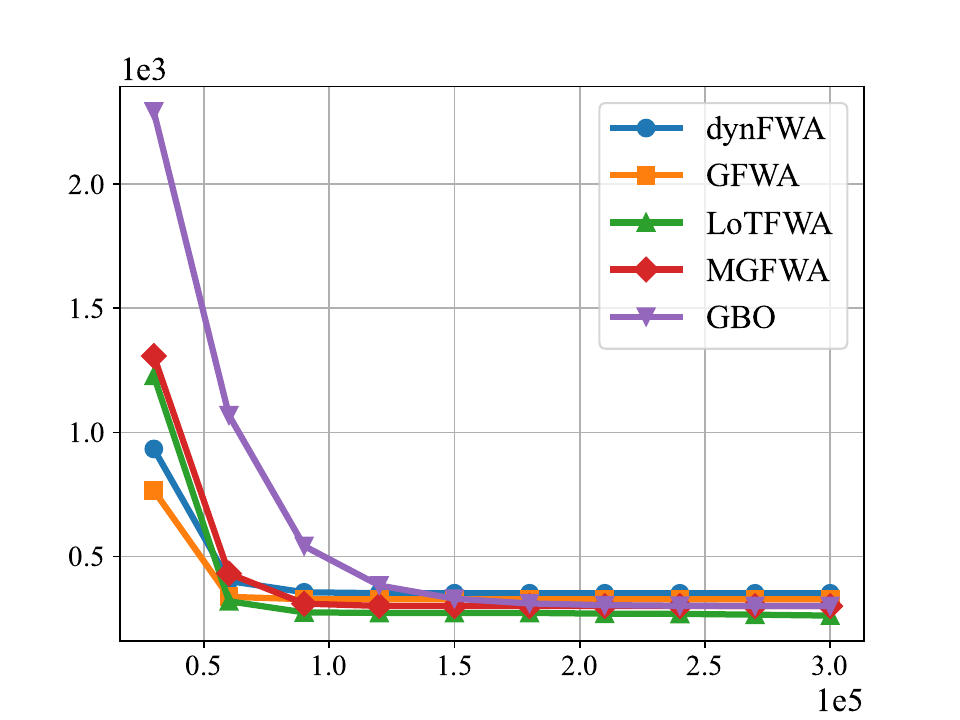}
    \caption{Convergence curves of different FWA algorithms on CEC2013.
    (Functions 1 to 28, arranged from left to right and from top to bottom.)}
        \label{cec2013_1}
\end{figure*}

Studying the characteristics of an algorithm's convergence curve can provide deeper insights into its performance and behavior. Since GBO and FWA share similar geometric properties in terms of population individuals, we plotted the convergence curves of GBO, dynFWA, GFWA, LotFWA, and MGFWA on the CEC2013 benchmark, as shown in Figure \ref{cec2013_1}. It can be observed that, in the early stages, GBO's convergence speed is slower than that of the other algorithms. However, in the later stages, GBO is less likely to get trapped in local optima and can converge to a better solution, which is related to the nature of GBO. This is because the search radius of the granular-balls decreases globally from large to small, and No-overlaping child granular-balls generation strategy prevents GBO from focusing on local optima in the early stages. According to the ``No Free Lunch" theorem, when comparing with FWAs under fixed resources, GBO is more effective in avoiding local optima, which accounts for its slower convergence speed in the initial phase compared to other algorithms.

\subsection{Figures and Tables}

\begin{table*}[htbp]
    \centering
    \renewcommand{\arraystretch}{1.2} 
    \caption{The parameter setting of comparison algorithms.}
    \label{table2}
    \begin{small}
    \begin{tabular}{lccc}
    \toprule
    Algorithms & Parameters & Values \\
    \midrule
    PSO & $N$,$c_1$, $c_2$, $w$ & 100,2, 2, 0.9-0.4 \\
    DE & $N$, $F$, $CR$ & 100, 0.5, 0.9 \\
    GA & $N$, $MR$, $CR$ & 100, 0.1, 0.8 \\
    ABC & $N$, $Limit$, $sn$ & 100, 200, 1 \\
    SHADE & $N$, $H$, $F$, $CR$ & 100, 100, 0.5, 0.5 \\
    LoTFWA & $fw_{size}$, $sp_{size}$, $init_{amp}$, $gm_{ratio}$ & 5, 300, 200, 0.2 \\
    JADE & $N$,$F$, $CR$, $pt$, $ap$ & 100,0.5, 0.5, 0.1, 0.1 \\
    MGFWA & $fw_{size}$, $sp_{size}$, $init_{amp}$, $gm_{ratio}$, $parameter_{N}$, $parameter_{b}$ & 5, 300, 200, 0.2, 10, 1.5 \\
    NSHADE & $N$, $F$, $CR$ & 100, 0.5, 0.5 \\
    LSHADE & $N$, $F$, $CR$ & 100, 0.5, 0.5 \\
    PVADE & $N$ & 100 \\
    SPSO2011 & $N$, $w$, $c_1$, $c_2$ & 100, $\frac{1}{2 \times \ln2}$, $0.5+\ln2$, $0.5+\ln2$ \\
    \bottomrule
    \end{tabular}
    \end{small}
    \vspace{-10pt}
\end{table*}

\end{document}